\documentclass{article}
\usepackage[utf8]{inputenc}
\usepackage{graphicx}
\graphicspath{{./image/}}
\usepackage{caption}
\usepackage{subcaption}
\usepackage{authblk}
\usepackage{lipsum}
\usepackage[dvipsnames]{xcolor}
\usepackage{bibentry}
\usepackage{bbm}
\usepackage{natbib}
\usepackage{tikz}
\usepackage{algorithm}
\usepackage{algpseudocode}
\usepackage{multirow}
\usepackage[table]{xcolor}
\usepackage{amsthm,amssymb,amsmath}
\usepackage{setspace}
\usepackage{booktabs}
\usepackage{float}
\usepackage{comment}
\usepackage{hyperref}

\newtheoremstyle{theoremsstyle}
  {15pt}   % ABOVESPACE
  {15pt}   % BELOWSPACE
  {\itshape\setstretch{1.15}}  % BODYFONT
  {}       % INDENT (empty value is the same as 0pt)
  {\bfseries} % HEADFONT
  {.}         % HEADPUNCT
  {5pt plus 1pt minus 1pt} % HEADSPACE
  {}       
\theoremstyle{theoremsstyle}

\author[1]{Ferdinand Bhavsar}
\author[1]{Lionel Benoit}
\author[2]{Maxime Savatier}
\author[1]{Edith Gabriel}
\affil[1]{Biostatistics and Spatial Processes (BioSP), INRAE, Avignon 84914, France}
\affil[2]{Scientific and Technical Department, ANDRA, Bure 55290, France}

\usepackage[a4paper,
            bindingoffset=0.2in,
            left=1in,
            right=1in,
            top=1in,
            bottom=1in,
            footskip=.25in]{geometry}

\title{Transformer-based Diffusion models for Hydrological Time Series Probabilistic Imputation and Forecasting}
\date{\today}

\begin{document}

\maketitle

\begin{abstract}
The modeling of hydrometeorological time series with limited observations is a key challenge in the monitoring of hydro-systems and water resources, as well as for flood or drought risk assessment. Due to the high variability of the underlying processes and the sparsity of available measurements, traditional statistical approaches often struggle to accurately represent their dynamics. In this context, recent advances in deep learning offer a promising direction for improving the representation and generation of complex temporal processes sampled at several observation sites. 

This study investigates the application of transformer-based diffusion models to the simulation and reconstruction of hydrological time series. The proposed framework is applied to the joint modeling of water quantity and quality at six sites spread across three adjacent headwater catchments located in North-East France on a limestone plateau covered by forests and field crops. The model is calibrated and validated using available observational data, which has been quality controlled and corrected for sensor drift and malfunction through collaborative efforts by LNE metrology expertise and Andra monthly quality control over more than 15 years. Its performance is compared with several established baseline approaches commonly used for time series modeling. Quantitative evaluation metrics are employed to assess the ability of the proposed method to reproduce key temporal characteristics of the observed signals in two settings: the imputation of incomplete time series and the forecasting of upcoming hydrological conditions. 

Results support the effectiveness of the transformer-based approach and highlight its capacity to capture and simulate the complex patterns present in hydrological data. In particular, the results indicate that diffusion models can efficiently sample realistic time series distributions under observation settings with variable missing data for both forecasting and imputation.
\end{abstract}

\section{Introduction}

In hydrological sciences, there is a growing need for the joint modeling of the quantity of water in rivers (i.e., streamflow) and the quality of this water (i.e., chemical, physical, and sometimes biotic signature). This applies in particular in the field of ecohydrology where the quantity and quality of water jointly impact stream ecosystems and stream-dependent organisms \citep{TEURLINCX201921}, and in the field of catchment hydrology where physical and chemical tracers (e.g., isotopic signature) are used to identify the origin, storage, and path of water through different compartments of the catchment \citep{hess-24-827-2020}. 

Water quality and quantity are increasingly monitored by automatic and multi-sensor stream gauges, creating a growing database of multivariate time series often recorded at several locations of the river network \citep{von2017lab, moiroux2023connecsens}. Such datasets often include gaps caused by sensor malfunctions, as well as heterogeneous start and end dates due to diverse station setup and discontinuation. This makes hydrological time series heterogeneous, and calls for methods of data augmentation (e.g. gap ﬁlling, spatial interpolation), either before integration in process-based models, or for other purposes linked to water management (decision support, early warning, digital twins, etc). Indeed, process-based models usually require structured input information representative of the different water storages and fluxes throughout the catchment of interest to properly depict streamflow generation processes, and available measurements rarely allow to fully quantify them \citep{data_augmentation_one, data_augmentation_zero}.

Data-based approaches are particularly well suited for data augmentation, and also allow for data emulation (e.g., stochastic weather \citep{obakrim2025multivariate} or streamflow generators \citep{hess-23-3175-2019}) as well as now-casting. We will therefore focus on this class of models, and more specifically on ensemble (a.k.a. probabilistic, generative) approaches allowing for the simulation of large ensembles of equally likely outcomes that can be used to assess the variability of the process at hand and to quantify the uncertainty of the simulations. Over the years, powerful parametric probabilistic models based on assumptions about the underlying probabilistic structure of the data have been developed. These models include Markov chains \citep{gmd-12-2767-2019}, Autoregressive Integrated Moving Average (ARIMA), \citep{ar_timeseries_models}, and Gaussian random fields \citep{10.1111/1467-9876.00419}. The main limitation of such models is their dependence on a prior parameterization that restricts their flexibility and, thereby, their ability to capture complex real-world phenomena realistically. To avoid this pitfall, non-parametric resampling methods get rid of explicit assumptions about the underlying distribution, and rely instead on the stochastic resampling of a training dataset. For example, Multi-Points statistics (MPS) simulation is a sequential simulation technique based on pattern matching \citep{mps_guardiano1993multivariate, mps_strebelle2000sequential, mariethoz2010direct}. 

Advances in deep learning have recently drawn considerable attention because these methods effectively capture complex and nonlinear relationships in a wide range of applications \citep{cybenko1989approximation}, and because they can be trained efficiently and flexibly when large quantities of data are available \citep{GoodBengCour16}. Over the past few years, long short-term memory (LSTM) models \citep{lstm_article} have been extensively explored for hydrological time series modeling, including the imputation of missing values \citep{lstm_gap_filling, lstm_gapfill_w} and forecasting \citep{Feng_2020, hess-22-6005-2018}. While they are theoretically able to model some long-range temporal dependencies (year long dependencies in hydrological time series), in practice LSTM models process time series sequentially and therefore struggle to capture complex global interactions. In contrast, the more recent transformer-based models leverage self-attention to model long-term dependencies more efficiently and in parallel \citep{vaswani2023attentionneed}. In hydrology, \cite{HU2024102695} used a transformer-based model for predictive forecasting of dissolved oxygen in rivers with greatly improved performance over baselines.

The deep-learning methods discussed above do not explicitly account for uncertainty, which is nevertheless essential in hydrology because the complexity of hydrological processes and the limited observations introduce substantial uncertainty in time series modeling. Rather than relying on deterministic models that attempt to recover a single estimate of the real time series, it is therefore preferable to endorse probabilistic methods that model uncertainty explicitly. In the field of deep learning, this is the approach followed by deep generative models, which are a new alternative class of simulation methods that sample from a simple and known latent distribution, and learn through optimization how to transform this latent distribution into a complex and unknown target distribution. Deep generative models have gained significant traction in the last few years for their outstanding performance. They learn to reproduce the complexity of the phenomenon using a training dataset large enough to be representative of its variability and complexity, and exhibit great results in learning to generate synthetic (i.e., never observed) realizations of complex multivariate distributions hardly distinguishable from observations \citep{arjovsky2017towards}. Deep generative models have been applied in generating realistic pictures such as face images \citep{brock_large_2019, zhu_unpaired_2020} and art images \citep{karras2019style}. For a long time Generative Adversarial Networks (GANs) \citep{goodfellow2014generative} have been the leading generative models, and have been successfully applied in hydrology; for example, \citet{chen_novel_2025} designed and trained a custom GAN model for flood forecasting. However, the practical application of GANs often suffers from stability issues, and training them can be a challenge \citep{goodfellow2014generative, arjovsky2017wasserstein, arjovsky2017towards, BHAVSAR2024105638}.

Due to these drawbacks, GANs have recently been overshadowed by denoising diffusion generative models \citep{sohldickstein2015deep, ddpm_ho}. Owing to their greater training stability and improved performance compared to GANs, these models have emerged as the state-of-the-art across a wide range of applications, including image generation and manipulation \citep{Gu_2022_CVPR, Kawar_2023_CVPR, Saharia_palette, Zhang_2023_ICCV, Ruiz_2023_CVPR, diffusion_beats_gans}, time series generation \citep{sound_diff, CSDI_NEURIPS2021, ecg_diffusion_beding_2025} and time series anomaly detection \citep{yang2023ddmtdenoisingdiffusionmask}. Diffusion models have also attracted increasing attention in geosciences, including hydrology. For instance, \citet{10.2166/wpt.2025.133} compared multiple models for dam inflow forecasting. The authors found that a hybrid architecture transformer-LSTM  performed better than diffusion and transformers architectures alone. Meanwhile, \citet{Ou_2025} designed a diffusion architecture with a Transformer backbone that outperformed LSTM baselines for forecasting using meteorological forecasts as covariates, the imputation of missing values being performed beforehand using a reanalysis dataset. In \citet{10777578} the authors used a diffusion model with a Transformer backbone for precipitation nowcasting, outperforming the performances of U-NET architectures. 

It appears that most existing hydrology-oriented diffusion approaches are designed for specific tasks, typically focusing on either forecasting or data imputation in isolation. This task-specific design limits the flexibility and in turn the broader potential of diffusion models, which are inherently capable of jointly addressing multiple objectives within a unified probabilistic framework, including generation, forecasting, and missing-value imputation. In contrast to hydrology, multi-task diffusion-based models have been successfully applied to the modeling of electrocardiogram time series \citep{alcaraz2023diffusionbasedtimeseriesimputation}, stock market and electricity production data \citep{yuan2024diffusiontsinterpretablediffusiongeneral}, as well as air quality and traffic time series \citep{CSDI_NEURIPS2021, 10.1145/3580305.3599257}.

In the present study we propose a modified diffusion model based on an existing transformer architecture to tackle both imputation and forecasting with a single model for applications in hydrology. We find that the resulting model is competitive, achieving lower error than multiple baselines on two datasets

The remainder of this document is organized as follows. First, we introduce some background on diffusion models and, more specifically, on conditional score-based diffusion models. We then show what modifications have been implemented to improve the architecture of the model, explaining our choices. Next, we present results on forecasting and imputation tasks on two datasets, one real and one synthetic, and compare these results to existing baselines. Finally, we provide some conclusions about the application of transformer-based diffusion models to hydrological time series probabilistic imputation and forecasting, and we propose some lines for future research.

\section{Background}

\subsection{Example dataset}
\label{subsec_exampledataset}
%\textbf{Cigéo Water Quality time series dataset:}

\begin{figure}[H]
    \centering
    \includegraphics[width=\linewidth]{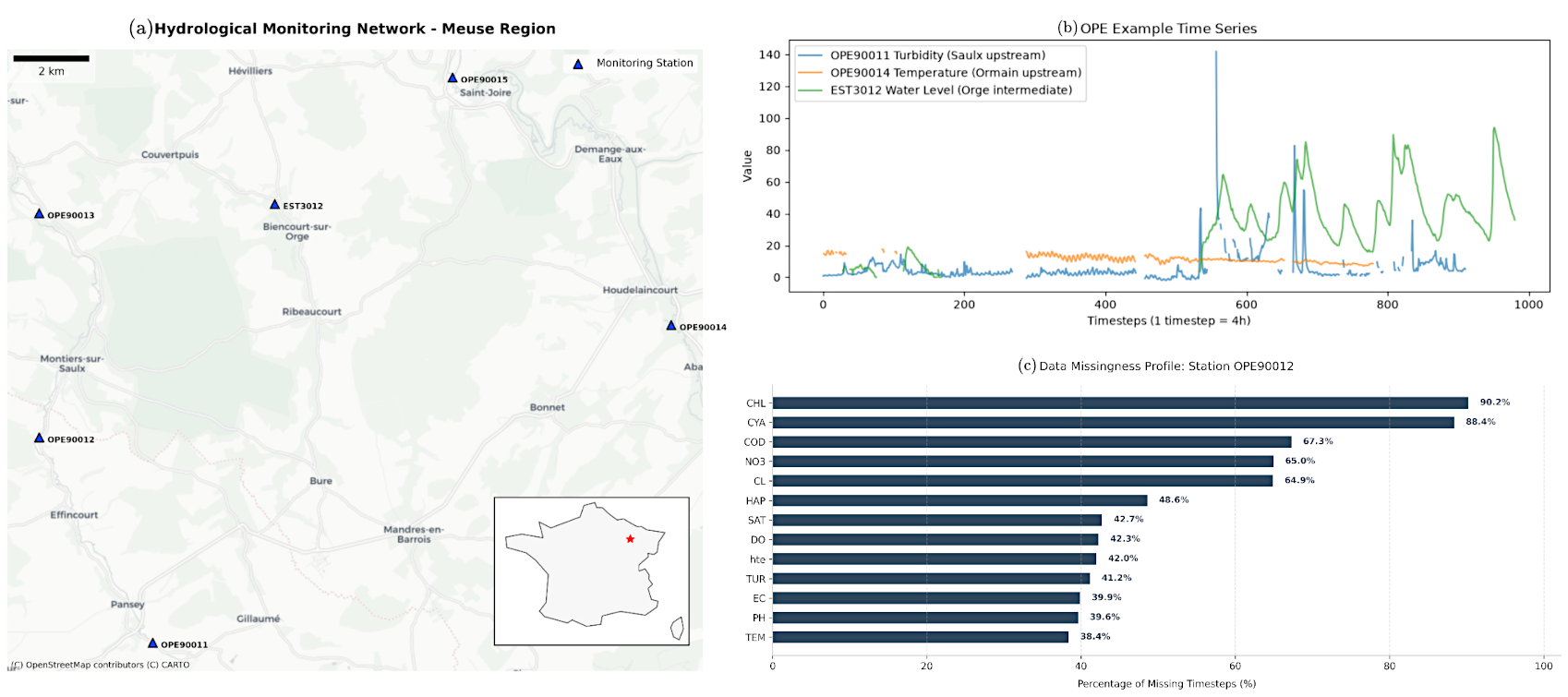}
    \caption{OPE hydrological monitoring network (Marne catchment - North-East France):  (a) Site description (blue triangles denote stream gauge locations); (b) Multivariate time series profile for some recorded features across 500 timesteps, including data gaps; (c) Percentage of missing data per variable at station 0PE90012.}
    \label{fig:ope_timeseries_profile}
\end{figure}

This study addresses the challenge of modeling hydrological time series within the framework of the Perennial Observatory of the Environment (in French Observatoire Pérenne de l'Environnement - OPE), a long-term research infrastructure dedicated to monitoring environmental and climate changes around the future Cigéo deep geological repository for radioactive waste. Data are collected from six monitoring stations located along the Orge, Saulx, and Ormain rivers, with a temporal resolution of 4 hours, resulting in an extensive dataset for analysis. Figure \ref{fig:ope_timeseries_profile} gives an overview of the dataset, which consists of water-quality time series covering an observation period from 2012 to 2024, i.e., 25980 timesteps. Each station records 13 variables. Among these, 10 were measured for the entire period: fluorescent dissolved organic carbon (fDOC, also referred to as fDOM, or fluorescent dissolved organic matter), dissolved oxygen in mg/L (DO), oxygen saturation in \% (SAT), electrical conductivity at 25°C (EC), Polycyclic aromatic hydrocarbons (PAH), nitrate concentration (NO3), water pH (pH), water temperature (TEM), turbidity (TUR), and water level (HTE). Additionally, 3 variables were only measured during the specific years of the measurements: Cyanobacteria (CYA), chlorophyll-a (CHL), dissolved chlorides (CL). The time series exhibit heterogeneous magnitudes: for example, pH remains within a narrow near-neutral range (mean=7.90, max=9.65), while turbidity spans several orders of magnitude (mean=22.52, max=3000.00). Data was corrected for bias by the French National Laboratory of Metrology and Testing (LNE) using monthly quality control by the OPE technicians and contractors \citep{Guigues_inpress}.

Some sensor (such as Chl-a, CYA, figure 1) were only deployed during the first years of the station life and for some stations the measurement timesteps were modified across the station life to ensure sufficient power stability on the stations in the early deployment periods between 2012 and 2017. Furthermore, frequent droughts and occasional sensor failure led to added missing data. Finally, in some cases for the period before 2017, sensors timesteps were not always synchronized between station. As a result, the proportion of missing data calculated in this work (which does not account for timesteps changes) can be substantial for some sensors, ranging from 25\% to 90\%.  It should be noted, that the high proportion of missing values reported for the OPE900XX stations is not fully representative of the actual sensor downtime or of the fraction of the year effectively covered by measurements. Rather, it also results from the non-homogeneous acquisition timestep of the raw data (ranging from 4 h to 8 h) which inflate the apparent missing-data rate when computed as a simple ratio of populated rows to expected time steps. This significantly increases the difficulty of modeling the hydrological and biogeochemical dynamics of the site. Another key characteristic of the dataset is its complexity: it exhibits temporal correlations across multiple time scales, cross-correlations between variables, and non-Gaussian distributions of most target variables.

In this context, two main tasks, imputation and forecasting, will be considered as goals for the study. Imputation aims to reconstruct missing observations in time series, providing experts with indicative information for more robust hydrological modeling, in particular when models require data with constant time steps and without gaps, which can be most challenging to have for long term measurements in rural or remote rivers and streams, especially when using self-powered stations (solar or battery powered). Forecasting helps to anticipate upcoming hydrological conditions, for instance accurate hydrometeorological forecasts can help predict floods, and better assess the transport and dispersion of pollutants linked to those episodes. %Therefore, deep learning methods are well-suited to extract and learn these complex patterns from imperfect data. %In particular, Transformer-based models are designed to capture intricate dependencies across both temporal and feature dimensions. However, they typically require large amounts of data to generalize effectively. Given the limited size of the dataset, this necessitates architectural adaptations to improve data efficiency and model robustness.

\subsection{Diffusion Models}

\begin{figure}[htbp]
    \centering
    \includegraphics[width=0.6\textwidth]{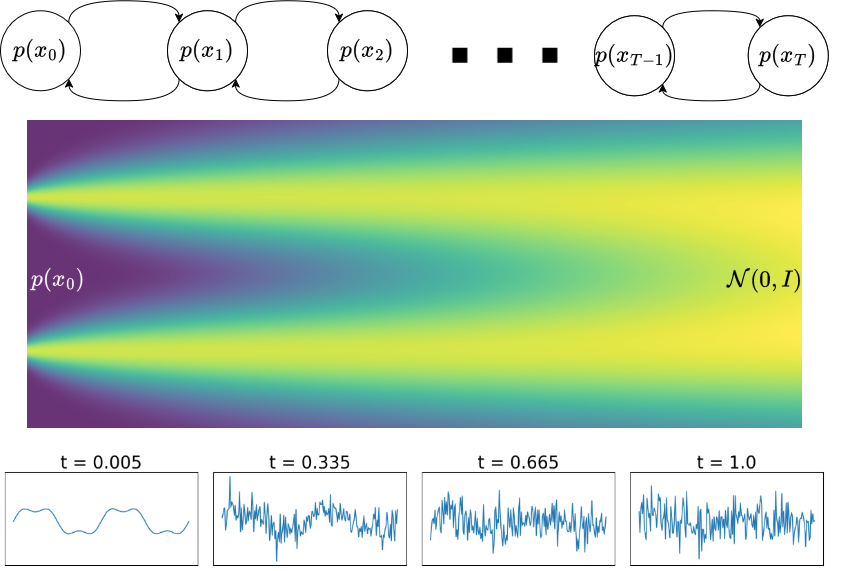}
    \caption{Illustration of the diffusion process, where $t \in \{0, 1, \dots, T\}$ indexes the diffusion timestep, with $t=0$ corresponding to the original data distribution and $t=T$ to the fully noised (Gaussian) distribution.
    \textbf{Top row}: marginal distributions $p(x_0), p(x_1), \dots, p(x_T)$ evolving over time.
    \textbf{Middle row}: Simplified representation of a data distribution $p(x_0)$ that starts with two separates modes representing its structure, and which is slowly noised until it becomes a standard Gaussian distribution ($p(x_T)$).
    \textbf{Bottom row}: example trajectory of a time series at different noise levels, illustrating the iterative noising process.
    }
    \label{fig:diffusion_process}
\end{figure}

Over the past few years, diffusion models have become one of the dominant generative modeling frameworks, in image synthesis and are attracting increasing attention for probabilistic forecasting and data imputation in scientific time series applications, including hydrology. Previous generative methods learned the parameters of a single transformation linking an easy-to-sample distribution to the target distribution. Variational Auto-Encoders (VAEs) learn this transformation using variational inference, mapping the data into a latent distribution and then back to the data space, thus learning the relation between the latent and data distributions \citep{kingma2014autoencoding}. Generative Adversarial Networks (GANs) learn the transformation by training a generator model to generate realistic data and a discriminator model to distinguish generated from real data, with both models trained alternately in an adversarial game.

In contrast to the above models involving a single transformation, diffusion models, inspired by non-equilibrium thermodynamics, generate data by learning an iterative denoising process that transform a simple, easy to sample noise distribution into the data distribution. Diffusion models have supplanted GANs and VAEs due to their performance in learning complex distributions and because their training is stable, i.e., less prone to issues such as mode collapse, adversarial optimization instabilities, and highly sensitive convergence dynamics \citep{sohldickstein2015deep, ddpm_ho, diffusion_beats_gans}.  %si possible spécifier ici ce que tu qualifie de "stable training"
 
%In this work, we want to sample new multivariate time series that describe water level and water quality, taking into account the natural variability and possible evolutions, with the added difficulty of missing values due to sensor problems.

%Over the past few years, generative models have gained fame as a class of deep learning models that are capable of learning probability distributions.
%In the past, they had great success for generating new, never-seen data such as natural images.
%The main generative models used today are Variational Auto-Encoders (VAEs), Generative Adversarial Networks (GANs) and Denoising Diffusion Generative Models (DDGM).
%Generally, we use these models when we want to sample new data points from the same distribution as a given dataset. 
%We focus our attention on DDGM, because of their performances and training stability. 

%These models often offer results equivalent or better than GANs, while being more stable during training. 
Denoising diffusion works by approximating the reverse of a noising process. This noising process is usually Gaussian, and therefore a Gaussian noise is added following a given schedule, until the noisy data approximately follow an isotropic
Gaussian distribution. This process is called the forward process, and is detailed hereafter.\\

Let $x_0 \in \mathbb{R}^n$ be a data point sampled from a real, unknown distribution $q$. A Gaussian noise $\epsilon_t$ with covariance matrix $\beta_t I_n$, where $I_n$ denotes the $n \times n$ identity matrix, is iteratively added to $x_0$ at step $t$ during $T$ evenly-spaced steps, with $\beta_t \in [0,1]$, $t \in \{0, 1, \ldots, T\}$, $T$ bounded. These noising steps define the following forward process:
\begin{equation}\label{eq:forward_process}
    x_t = \sqrt{1-\beta_t}x_{t-1} + \sqrt{\beta_t}\epsilon_t, \hspace{25pt} \epsilon_t \sim \mathcal{N}(0, I_n)
\end{equation}

where each step in the forward process depends solely on the previous one, thus defining a Gaussian Markov Chain:
\begin{equation}\label{eq:diffusion_process_init}
    p(x_{1:T}|x_0)=\prod^{T}_{t=1}p(x_t|x_{t-1}), \text{where } p(x_t|x_{t-1}) = \mathcal{N}\Big(\sqrt{1-\beta_t}x_{t-1}, \beta_tI_n\Big)
\end{equation}

where $\beta_t$ is arbitrarily fixed by defining the noising schedule. Denoising diffusion relies on the choice of a well-behaved noising schedule, and it is therefore necessary to choose an increasing variance schedule such that when $t \xrightarrow{} T$ the sample distribution is approximately the one of a white noise vector (i.e., $x_T \sim \mathcal{N}(0, I_n)$). In \citet{ddpm_ho}, the authors chose to have a linear schedule from $\beta_1 = 10^{-4}$ to $\beta_T = 0.02$, but \citet{improved_ddpm} introduced a cosine schedule which puts more focus on the beginning and on the end of the diffusion process, which yielded improved results.

From equation \ref{eq:diffusion_process_init} one can compute the marginal distribution $q(x_t|x_0)$ in its closed form and this reparameterization allows to find $x_t$ without applying the entire forward process:
\begin{equation}\label{eq:closed_forward_process}
    x_t = \sqrt{\Bar{\alpha_t}}x_0 + \sqrt{(1-\Bar{\alpha_t})}\epsilon,
\end{equation}
where $\alpha_t = 1 - \beta_t$, $\bar{\alpha_t} = \prod_{s=1}^t\alpha_s$, and $\epsilon$ is a standard normal variable (i.e., $\epsilon \sim \mathcal{N}(0, I_n)$).\\

It can be shown that a reverse process exists, and that it transforms the white noise at time $T$ into a data point at time $0$ by reversing the noising process through a sequence of denoising steps \citep{ddpm_ho}. Conditioned on $x_0$, this reverse process follows a Gaussian distribution:
\begin{equation}
p(x_{t-1} \mid x_t, x_0) = \mathcal{N}\big(\mu_t(x_t, x_0), \sigma_t^2 I_n\big).
\end{equation}

Therefore, if $x_0$ is known, the denoising step at time $t$ can be computed as:
\begin{equation}\label{eq:reverse_process_sample}
x_{t-1} = \mu_t(x_t, x_0) + \sigma_t \epsilon
\end{equation}

where $\epsilon$ is Gaussian white noise and $\sigma_t$ is a standard deviation that depends only on $\beta_t$. Using equation \ref{eq:reverse_process_sample} the reverse process can now be sampled if $\mu_t(x_t, x_0)$ can be computed. However, $x_0$ is not accessible during sampling, and an approximation has to be found instead. Using the reparameterization from \cite{ddpm_ho}, $\mu$ can be re-written in the following form:

\begin{equation}
    \mu_t(x_t, x_0) = \Tilde{\mu_t}\Bigg(x_t, \frac{1}{\sqrt{\alpha_t}}\Big(x_t - \sqrt{(1-\Bar{\alpha_t})}\epsilon\Big)\Bigg) = \frac{1}{\sqrt{\alpha_t}}\Big(x_t - \frac{\beta_t}{\sqrt{(1-\Bar{\alpha_t})}}\epsilon\Big)
\end{equation}

A neural network is trained to approximate the reverse process, which slowly removes the noise from the noisy data point until its original structure is recovered. This network (with weights $\theta$) takes as input the noisy time series $x_t$ and the step $t$:
\begin{equation}
    \mu_\theta(x_t, t) = \frac{1}{\sqrt{\alpha_t}}(x_t - \frac{\beta_t}{\sqrt{(1-\Bar{\alpha_t})}}\epsilon_\theta(x_t, t))
\end{equation}

To approximate the reverse distribution it can be shown that the neural network only needs to be trained to learn how to predict $\epsilon_\theta$, which is an approximation of the real noise $\epsilon$ added to the original data-point $x_0$. The following loss function is used for the training:
\begin{equation}
\begin{aligned}
\theta^* = 
\mathbb{E}_{x_0,\epsilon,\;t\neq 0}
\Bigg[
\frac{1}{2}
\frac{1-\bar{\alpha}_{t}}{1-\bar{\alpha}_{t-1}}
\beta_t
\left\|
\epsilon -
\epsilon_\theta\!\left(
\sqrt{\bar{\alpha}_t}x_0 +
\sqrt{1-\bar{\alpha}_t}\,\epsilon,
t
\right)
\right\|_2^2
\Bigg]
+ C.
\end{aligned}
\end{equation}

\cite{ddpm_ho} observed that the weighting terms have little impact on the quality of the generated samples. This leads to the following noise prediction loss:
\begin{equation}
\begin{aligned}
\theta^*
=
\operatorname*{argmin}_{\theta}
\mathbb{E}_{x_0,\epsilon,t}
\Bigg[
\left\|
\epsilon -
\epsilon_\theta
\left(
\sqrt{\bar{\alpha}_t}x_0+
\sqrt{1-\bar{\alpha}_t}\epsilon,t
\right)
\right\|_2^2
\Bigg],
\end{aligned}
\end{equation}

%A neural network is then trained to approximate this reverse process, which is a process that slowly removes the noise from the noised data point until we arrive to its original structure. The networks weights are noted $\theta$, and takes as input the noised image $x_t$ and the step $t$.

%\begin{equation}
%x_{t-1} = \mu_\theta(x_t, t) + \sigma_t \epsilon
%\end{equation}

%where $\mu_\theta(x_t, t)$ is computed using the neural network.\

Given the approximation, a new data point can be generated. Starting at time $T$, $x_T$ is sampled from a normal distribution. Then, working backward through time, the noise $z$ is repeatedly sampled and each previous state $x_{t-1}$ is calculated using the statistical functions learned by the model:

\begin{equation}
\begin{aligned}
z &\sim \mathcal{N}(0, I), \\
x_{t-1} &= \mu_\theta(x_t, t) + \sqrt{\Sigma_\theta(x_t, t)} \, z.
\end{aligned}
\end{equation}

The reverse process continues iteratively until it reaches $x_0$, the newly sampled data-point.\\

%Once the network is trained, it can be used to transform any noise to a new datapoint. This approximated reverse process can be described as an iterative refinement using a neural network.
Because the example dataset introduced in section~\ref{subsec_exampledataset} encompasses a large amount of missing data, we focus in the following on a diffusion model that can be trained while taking gaps into account.

\subsection{Conditional Score-based Diffusion Models}
\label{subsec_originalCSDI}

\paragraph{Conditional Score-based Diffusion Models (CSDI).}
Conditional Score-based Diffusion Models (CSDI) are a type of diffusion models specifically designed for time series with missing values \citep{CSDI_NEURIPS2021}. They are inspired by masked language modeling and are tailored to explicitly condition on observed values when predicting the missing ones. CSDI learns to fill in missing values in time series while respecting their temporal structure as well as relationships between variables.

In \citet{CSDI_NEURIPS2021} the authors designed both the architecture and the training paradigm of CSDI to handle incomplete time series. The model takes observed values as conditional inputs and predicts missing values accordingly. In the original paper, the authors applied CSDI to both imputation and forecasting tasks, using separate models for each task.

\paragraph{Problem setup.}
Given conditional observations $x^{co}_0 \in \mathbb{R}^d$, with $0\leq d<n$, and imputation targets $x^{ta}_0 \in {n-d}$, noisy targets are sampled as:
\begin{equation}
x^{ta}_t = \sqrt{\bar{\alpha}_t}\, x^{ta}_0 + \sqrt{1 - \bar{\alpha}_t}\, \epsilon,
\end{equation}
where $\epsilon \sim \mathcal{N}(0, I_n)$ and $\alpha_t$ follows a predefined noise schedule. The denoising network $\epsilon_\theta$ is trained by minimizing the following loss:
\begin{equation}
\min_{\theta} \mathcal{L}(\theta)
= \mathbb{E}_{x_0 \sim q(x_0),\, \epsilon \sim \mathcal{N}(0, I),\, t}
\left[
\left\| \epsilon - \epsilon_\theta\left(x^{ta}_t, t \mid x^{co}_0 \right) \right\|^2
\right].
\end{equation}

\paragraph{Transformers.} The backbone of the model is based on the Transformer architecture \citep{vaswani2023attentionneed}, which has demonstrated strong performance in modeling long-range dependencies through self-attention. Attention is a powerful operation that allows to model dependencies within sequential data, with no limitations regarding to their distance in the sequence \citep{Wangetal2014, Bahdanauetal2015}. It is defined as:

\begin{equation}
\begin{aligned}
\text{Attention}(Q, K, V) &= \text{softmax}\Bigg( \frac{Q K^\top}{\sqrt{d_k}} \Bigg) V, \\
Q = X W_Q, \quad K &= X W_K, \quad V = X W_V, \\
\end{aligned}
\end{equation}
where $W_Q, W_K, W_V \in \mathbb{R}^{d \times d_k}$ are matrices with trainable coefficients, and $X \in \mathbb{R}^{n \times d}$ is an input multivariate time series.

Unlike recurrent architectures, Transformers process the full sequence in parallel, which captures global temporal interactions efficiently. Given an input sequence representation $x \in \mathbb{R}^{T \times d}$, multi-head attention allows the model to attend to multiple representation subspaces simultaneously:

\begin{equation}
\text{MultiHead}(X) = \text{Concat}(\text{head}_1, \dots, \text{head}_H)W_O,
\end{equation}

where $\text{head}_i, 0\leq i\leq H$ is an independent attention operation with its own trainable set of weights:

\begin{equation}
\text{head}_h = \text{Attention}(Q^h, K^h, V^h)
\end{equation}

where $W_Q^h, W_K^h, W_V^h \in \mathbb{R}^{d \times d_k}$ and $W_O \in \mathbb{R}^{Hd_k \times d}$ are learned parameters.

After multi-head attention a two-layer fully connected network called a Multi-Layer Perceptron (MLP) is applied independently to each timestep. The MLP is a per-timestep nonlinear layer that converts the multi-head attention outputs, i.e. temporal and features context, into richer latent features and stores learned patterns in its weights.:

\begin{equation}
\text{MLP}(x) = \sigma(x W_1 + b_1) W_2 + b_2,
\end{equation}

where $\sigma(\cdot)$ denotes a non-linear activation function.

\paragraph{Model architecture.}
Unlike most diffusion models, CSDI does not rely on a U-Net architecture \citep{ronneberger2015unet}. Instead, it consists of a stack of layers grouped in blocks operating all at a fixed resolution throughout the network. Each block contains a skip connection, which allows the input of the block to bypass the intermediate layers and be added directly to the block’s output. In CSDI's architecture, blocks are therefore called residual blocks \citep{resnet_og_paper}.

The model takes two input channels. The first is the main input, defined as the concatenation of the noisy target values ($x^{ta}_t$) and the conditional observations ($x^{co}_0$):
\begin{equation}
x = \left[ x^{ta}_t,\; x^{co}_0 \right].
\end{equation}

The second input channel consists of side information, which includes supplementary data that can improve prediction quality, such as the observation mask and optional temporal or feature embeddings.

To capture temporal dependencies and inter-feature relationships, each residual block contains two Transformer layers: one operating along the temporal axis and another one operating along the feature axis. These two layers enable the model to learn complex dependencies across time and between variables.

At each generation step, a residual block produces two types of outputs: (1) a residual output which is the input of the next residual block, (2) a secondary output, which is stored (skip connection). The final output of the network uses all the skip connections that are aggregated and are then passed through final convolutional layers.

\paragraph{Training strategy.}
The training procedure of CSDI is inspired by masked language modeling and differs from that of standard diffusion models. The reverse denoising process cannot be directly approximated during the inference because the true target values $x^{ta}_0$ are unknown. Instead, during training, artificial missingness is introduced by masking observed values. Hence, given a sample $x_0$ from the dataset, the observed entries are randomly split into two disjoint subsets: one subset is treated as imputation targets $x^{ta}_0$, while the other subset is used as conditional observations $x^{co}_0$.

In the original CSDI paper, multiple masking strategies for training were proposed. The random strategy selects imputation targets by randomly masking a percentage of observed values, while the historical strategy leverages patterns from training data by aligning observed and missing indices between samples to capture structured missingness. The mixed strategy combines both approaches, improving generalization while learning realistic missing patterns.  In this work, a combination of random masking and forecasting masking strategies is employed. In the latter, future time steps are masked to train the model for prediction tasks. %In our work, we employ a mix of random masking strategy is employed, and forecasting masking, where future time steps are masked to train the model for prediction tasks.
%The authors argue that MAR masking yields better empirical performance compared to alternative masking schemes.

%which takes as input at each step white noise where the data are missing ($x_t^{ta}$, with $ta$ meaning target'') and the observed values ($x_0^{co}$, with $co$ meaning conditional observation''). 

%\subsection{Related Work}
\section{Methods}

\subsection{Augmented CSDI for hydrometeorological time series}
\label{subsec_augmentedCSDI}

Due to a limited training sample size, a high dimensionality and an heterogeneous scaling of the hydrological signals involved, the original CSDI architecture presented in section~\ref{subsec_originalCSDI} performed poorly when applied to the test dataset introduced in section~\ref{subsec_exampledataset}. The main challenges faced by the model were (see Appendix~\ref{sec:ablations_tests} for details):
\begin{itemize}
\item the training failed to converge to a proper solution due to training instability and poor convergence,
\item the simulated time series ignored the observed values, failing to capture local relationships and resulting in jagged artifacts,
\item the simulations collapsed toward a single mean, with individual time series showing little variation around it,
\item the diurnal (and multi-day) cycles were somewhat captured, but the timing of sporadic peaks often did not align with observations.

\end{itemize}

To overcome these limitations we designed the updated CSDI architecture detailed below.

\paragraph{Custom CSDI architecture}\label{sec:custom_model}

The modification of the original CSDI architecture to better comply with hydrological data followed a step by step approach, each step being guided by the analysis of the limitations listed above. See Appendix \ref{sec:arch_comparison} for a comparison of the two architectures.
  
To improve the simulation of local patterns during imputation and enhance the conditioning to observed values, multiple convolutional layers have been added and combined with a multi-scale layer in each residual block in the form of a Residual U-block (RSU), based on the nested U-shaped design introduced in U²-Net \citep{u2net}. Each RSU consists of a lightweight U-Net–style convolutional sub-network embedded within the residual mapping. The time series are down-sampled twice along the temporal axis and processed at progressively lower temporal resolutions before being up-sampled and fused back. Internal skip connections within the RSU preserve fine-grained temporal information while keeping broader contextual information across scales. This design allows each residual block to jointly model local and global temporal patterns in a parameter-efficient manner, thanks to the pooling layers.

To improve the inclusion of the side information (i.e., the timesteps encoding and the mask) to the simulations, we changed the manner in which this information is integrated into the model. The side information was revamped, with three objectives: (1) side information should integrate the observed values and (2) it should have increased control over the hidden time series states, while (3) the timesteps should encode periodic signals explicitly.

For the first objective, the observations $x_0^{co}$ were simply appended to the side information. Although these observations are already present in the residual stream, this results in the information now being input twice, with each pathway serving a distinct functional purpose. The residual branch integrates observed values through convolution and attention, capturing both local and long-range patterns. In contrast, the side channels leverage the observed values alongside the rest of the side information to modulate the residual stream in a complementary manner (see Figure \ref{fig:arch_comparison} for more details).

For the second objective, instead of inputting the data from the side information through a simple additive biasing, a modulation mechanism inspired by StyleGAN \citep{karras2019style} was adopted. Specifically, the model learns two conditioning-dependent components: a multiplicative scaling term and an additive offset, which together modulate the activations and enable more expressive and reliable use of the conditioning information. Given an input sequence representation $x \in \mathbb{R}^{T \times c}$ and the side information representation $y \in \mathbb{R}^{T \times d}$ the following conditioning is applied:

\begin{equation}
    \text{Conditioning}(x, y) = x \odot \alpha(y) + \beta(y) % TODO, placeholder
\end{equation}

where $\alpha \in \mathbb{R}^{T \times c}$ and $\beta \in \mathbb{R}^{T \times c}$ are the outputs of a convolution layer applied to the side information, and $\odot$ is the element-wise multiplication.\\

Finally, for the third objective, inspired by feature engineering from time series literature \citep{bansal2025temporalencodingstrategiesenergy}, we include explicit encoding of some cycles. Previous deep learning for hydrology literature recommends similar Fourier encoding to help capture seasonal pattern dependencies within the time series properties and enhance the model's performance \citep{he2025deeplearningfoundationpattern}. Given the timesteps $t \in \mathbb{N}$ of the time series, Fourier encoding is used to represent periodic structure at multiple scales. The Fourier timesteps mapping is defined as:

$$\Phi(t)=\big[\sin(\tfrac{2\pi}{\alpha_1}t),\cos(\tfrac{2\pi}{\alpha_1}t),\dots,\sin(\tfrac{2\pi}{\alpha_N}t),\cos(\tfrac{2\pi}{\alpha_N}t)\big]\in \mathbb{R}^{2N}$$

where $\{\alpha_i\}_{i=1}^N$ denote a set of predefined cycle lengths. In our use case, we use cycles of 1 day, 7 days,  30 days, 90 days and 365 days.

The resulting representation is fixed and augments the original temporal input with explicit phase-aware periodic features, enabling the model to capture both short- and long-term temporal dependencies associated with the specified cycles.\\

While the above architecture modifications led to improvements in simulated time series, it also led to instabilities during the training, with the gradients diverging to extremely high values. To alleviate this issue, Root Mean Squared Normalization (RMSNorm, \citet{rms_prop}) was applied in each residual block. RMSNorm focuses on re-scaling invariance and regularizes the summed inputs according to the root mean square (RMS) statistic:

\begin{equation}
\mathrm{RMSNorm}(x_i) = \frac{x_i}{\sqrt{\epsilon + \frac{1}{C}\sum^C_{i=1}x^2_i}} \odot \gamma_i,
\quad \text{for } i \in \{1, \dots, C\}
\end{equation}

where $\epsilon << 1$, $C\in \mathbb{N}$ is the number of channels in the residual block, and $\odot$ is the element wise product. This addition of RMSNorm has already been identified in the literature for stabilization and performance improvement \citep{jones2026elucidatingdesignspaceflow}, and is related to AdaLN in Diffusion Transformers (DiT, \citep{peebles2023scalablediffusionmodelstransformers}).

\paragraph{Covariate Augmented Custom CSDI model}\label{sec:method_covariates_model}

The custom architecture offers satisfactory results for imputation but its forecasting performance remains limited when unforeseen external events (e.g., a rainfall event generating a streamflow response, and in turn a change in the chemical composition of stream water) induce distributional shifts in the hydrological time series.

However, such events and in particular droughts and rains can be forecasted using numerical weather models, and the associated information will be leveraged in the Covariate Augmented Custom CSDI model by simulating the time series $x$ knowing the covariate data $y$ derived from numerical weather forecasts. This conditional distribution can be modeled directly in the generator network. This idea was first exploited in conditional GANs by \citet{mirza2014conditional} to generate according to some conditioning parameters. This has since been adapted to generate conditional images and text with denoising diffusion models, see e.g., \citet{saharia2022paletteimagetoimagediffusionmodels, Dieleman2022ContinuousDF}.

In practice, the denoising network is extended to incorporate covariates as additional inputs:
\begin{equation}
\hat{x}_0 = D_\theta(x_t, t, x^{co}_0, y),
\end{equation}
where $x_t$ denotes the noisy sample at diffusion step $t$, $x^{co}_0$ represents the observations, and $y$ is the covariate time series. The model implicitly learns the conditional distribution through the standard diffusion objective, without modifying the loss function. In other words, conditioning is achieved purely through the architectural augmentation.

%The overall covariate-augmented architecture remains largely identical to the custom architecture presented previously. The main modification consists of concatenating the covariates with the residual stream before each residual block. %Moreover, to mitigate the risk of over-reliance on covariates and improve generalization, we additionally introduce dropout layers within the residual blocks.

%In an hydrology use-case, many different covariates may be used to inform our neural network and lead to better forecasting results.

\subsection{Baselines}

To assess the performances of the custom architecture, we compare the new model against three types of baselines. First, we consider classic statistical methods, including Gaussian Processes and Multi-point Statistics. These methods are well established for capturing correlations in space, time and between variables, and also for quantifying uncertainty. Refer to Appendix \ref{app:gp} for more details on the implementation of the Gaussian process in the present setting.

Second, we compare the custom CSDI model to other neural networks dedicated to time series processing. These include the base CSDI architecture described in section~\ref{subsec_originalCSDI} as well as a U-NET architecture, which is commonly used in image diffusion models \citep{ronneberger2015unet, zhang2024emergencereproducibilitygeneralizabilitydiffusion}. 

Third, forecasting tests on a synthetic dataset include two additional forecasting baselines: TSMixer \citep{chen2023tsmixerallmlparchitecturetime} and NHits \citep{challu2022nhitsneuralhierarchicalinterpolation}. These models are not evaluated on imputation tasks because they do not handle missing values. While NHits is deterministic, TSMixer has a probabilistic and deterministic version. We use both versions as baselines, naming them respectively TSMixer-P and TSMixer-D. This allows us to compare both point-wise prediction models and probabilistic baseline neural-networks. We use the Darts implementation for these last two baselines \citep{JMLR:v23:21-1177}.

\section{Results}\label{sec:results}

%\subsection{Datasets}\label{sec:datasets}

% We make a random syntetic dataset that seeks to mimic
% Consist of 8 concatenated multivariate time series, with three components, one dynamic, one seasonal correlated to the dynamic, one independant seasonal. Finally, a univariate timeserie with regime change to model a station with regular droughts.
% A more detailed description is available in annex

\subsection{Evaluation settings}\label{sec:eval_settings}
All experiments were conducted on an NVIDIA RTX 2000 Ada Generation GPU equipped with 16 GB of RAM.

To evaluate our model, we use two datasets: (1) the OPE dataset introduced in section~\ref{subsec_exampledataset}, and (2) a synthetic time series dataset that seeks to mimic hydrological dynamics, which has been designed to test how the models perform on a complete dataset and how they adapt to new data. 

In all cases all variables were normalized using a standard normalization procedure. For the covariates-augmented model, we use weather covariates derived from the SAFRAN reanalysis \citep{quintana2008analysis}. For each OPE station the covariates time series were extracted at the closest 8km grid point in the SAFRAN dataset. The resulting series are normalized and passed to the model as a secondary input as described in Section \ref{sec:method_covariates_model}.\\

\paragraph{Synthetic time series dataset:} Inspired by \citet{su172210295}, we generated a synthetic dataset that seeks to mimic the hydrological time series of the example dataset. The synthetic dataset consists of 8 concatenated multivariate time series, with a timestep of 1h, and with three components for each multivariate time series: (1) a dynamic component, (2) a periodic component correlated with the dynamic one that comprises two periods of approximately 24h and 30 days, and (3) a periodic component independent of the others with periods of approximately 48h and 25 days. Finally, a univariate time series with regime switches is used to model a station experiencing droughts. A more detailed description is available in Appendix \ref{annex:synthetic}. The performance on the synthetic dataset was evaluated using all models. In contrast, the performance on the real-world dataset was assessed only for the models capable of both imputation and forecasting because this dataset encompasses missing values.\\

\begin{figure}[t]
    \centering
    \includegraphics[width=\linewidth]{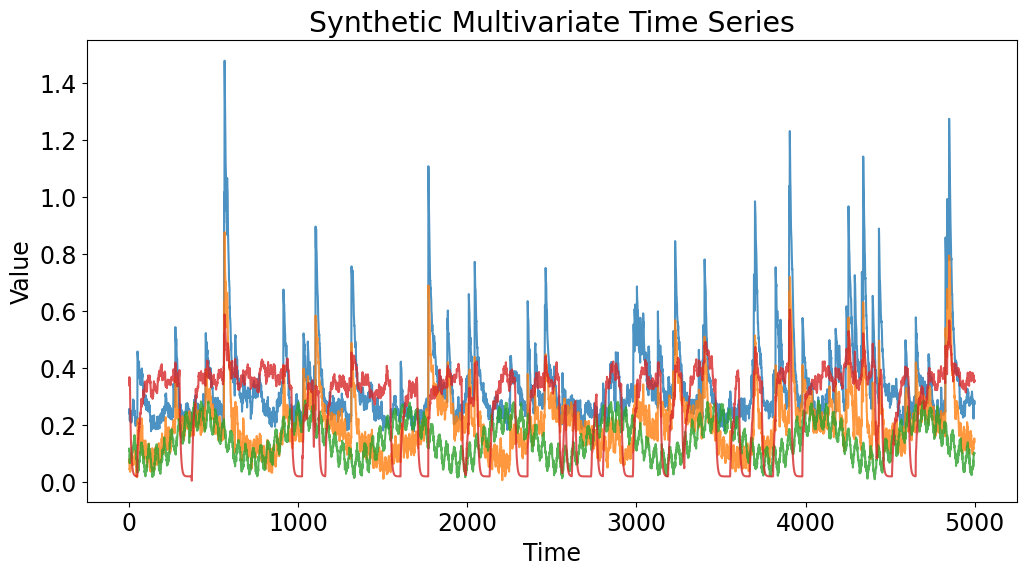}
    \caption{Synthetic data generated to test models. In blue we have a dynamic component, in orange a periodic component linked to the dynamic component, in green an independent periodic component, and in red a single time series with drought like periods (see Annex \ref{annex:synthetic} for more details).}
    \label{fig:synthetic_data_}
\end{figure}

\paragraph{Evaluation metrics:} We adopt both deterministic (Root Mean Squared Error, Mean Absolute Error, SMAPE) and probabilistic (CRPS) metrics for model evaluation.  The deterministic metrics measure the point-wise accuracy of the models, while the CRPS evaluates the fit of an estimated probability distribution to a ground-truth observation.

\paragraph{Test Sets Configuration:} For imputation ground truth, we randomly remove 10\%, 25\%, 50\% and 90\% values on 24 timesteps test time series. The points to mask were all selected with the same seed for every model, so that the comparison was based on the same set of observed values. A different seed was used for each hydrological variables, meaning that the missing steps varied along the last axis.
For forecasting, we remove the 6, 9, 12 last timesteps of 24 timesteps test time series. Evaluations are computed between the masked ground truth and the median of the simulations for probabilistic models, or the predicted timeserie for predictive models. See Appendix \ref{fig:eval_framework} for an overview of the evaluation framework.

\begin{figure}[H]
    \centering
    \includegraphics[width=\linewidth]{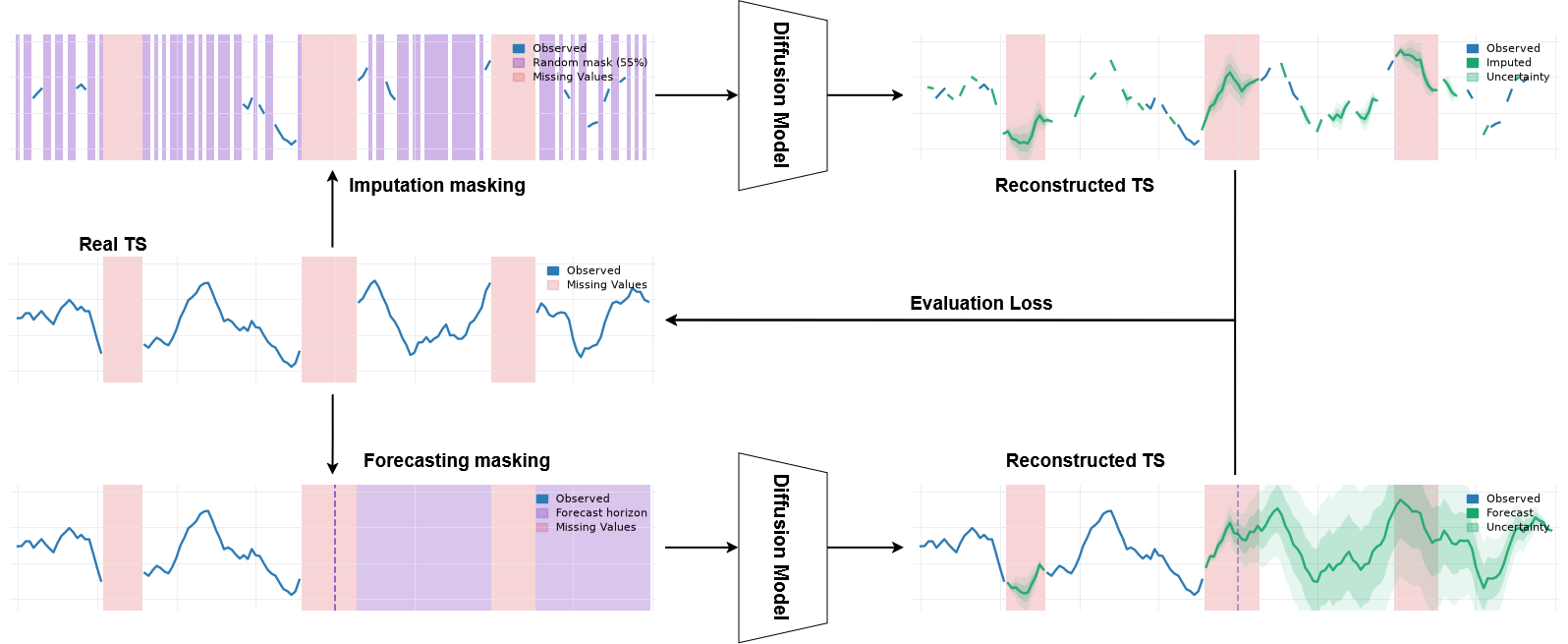}
    \caption{The evaluation setting of our diffusion model for imputation (top) and forecasting (bottom).}
    \label{fig:eval_framework}
\end{figure}
 
\subsection{Performance on the synthetic dataset}

\begin{figure}[H]
% ===================== FIGURES =====================
\begin{minipage}{0.5\linewidth}
\centering
\includegraphics[width=\linewidth]{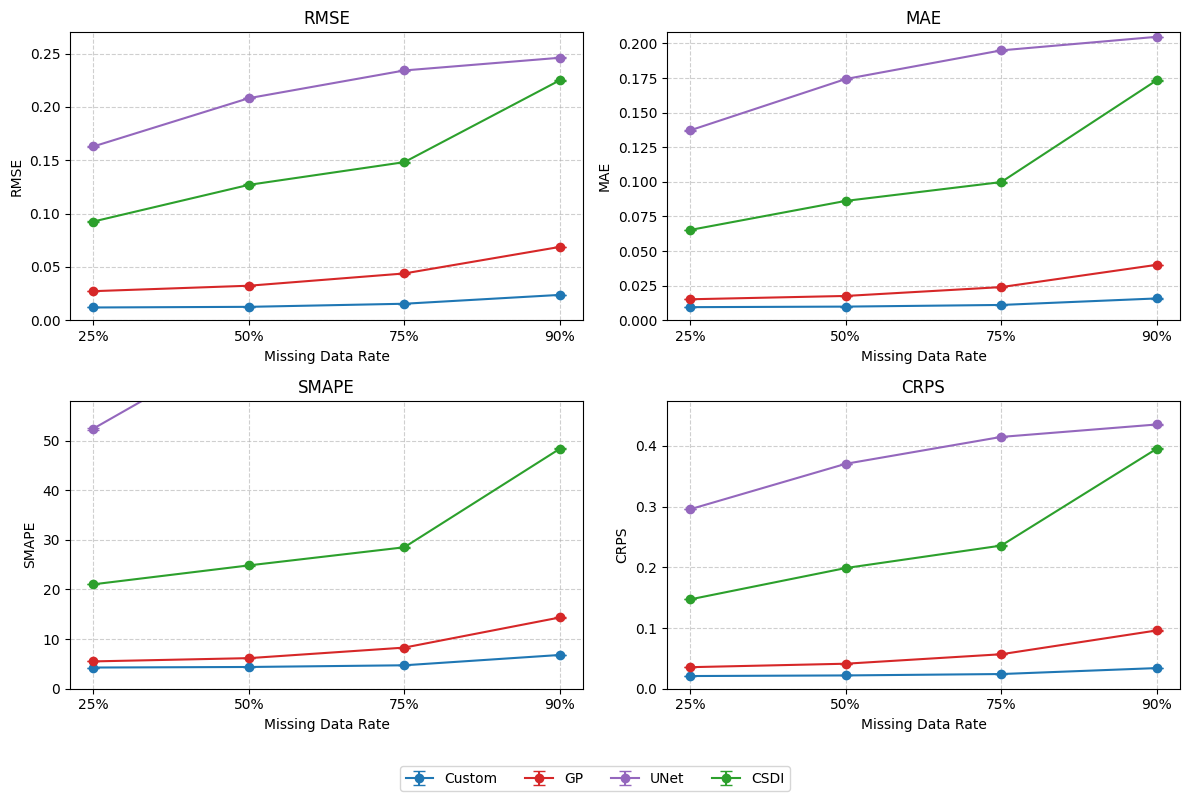}
\caption*{(a) Imputation Metrics Visualization}
\end{minipage}
\hfill
\begin{minipage}{0.5\linewidth}
\centering
\includegraphics[width=\linewidth]{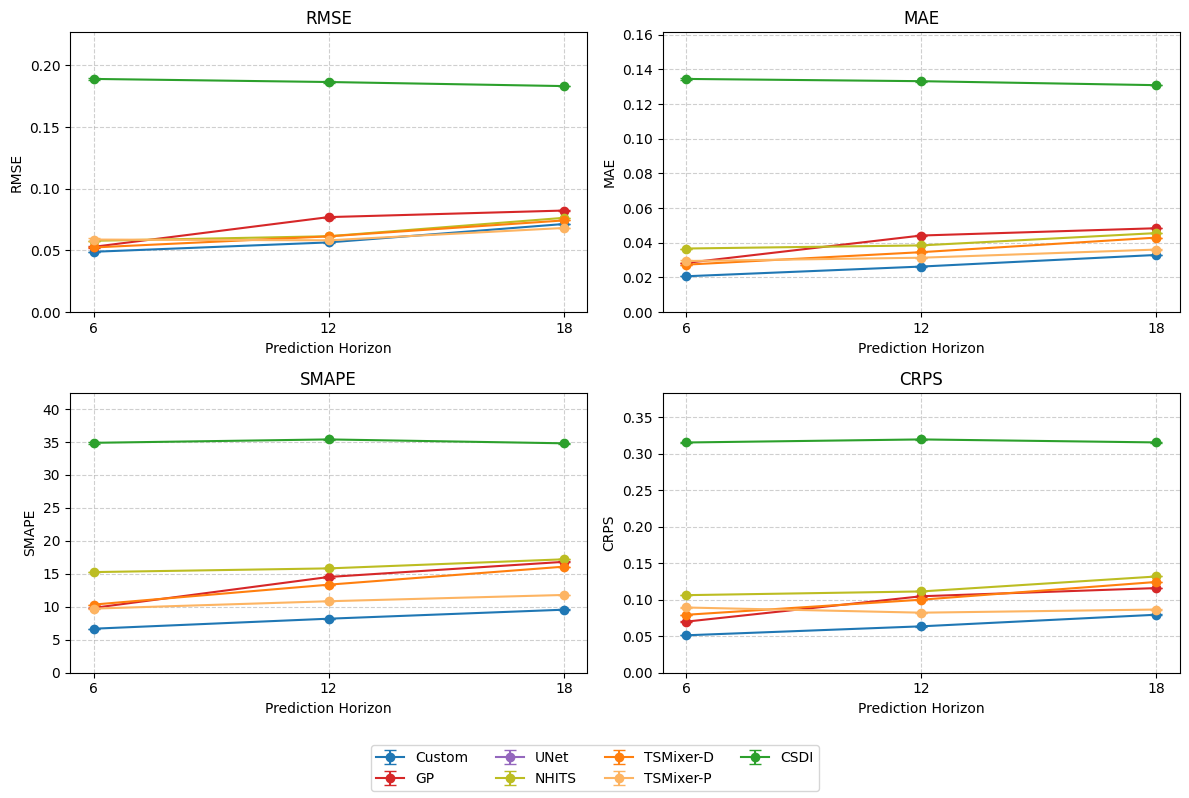}
\caption*{(b) Forecasting Metrics Visualization}
\end{minipage}
\caption{
Comparison of model performance on the synthetic dataset across missing ratios and prediction horizons. For clarity the metrics of the U-NET model are omitted or left out of bounds when it substantially underperforms compared to other models.
}
\label{fig:syn_data_res_graph}
\end{figure}

The models are trained and tested on the synthetic dataset described in Section \ref{sec:eval_settings}. Figure \ref{fig:syn_data_res_graph} displays the performance of the different models for the two tasks of interest, namely imputation and forecasting.

Results in Figure~\ref{fig:syn_data_res_graph}.a show that for imputation all the metrics support that the custom CSDI model outperforms the three baselines. The GP benchmark leads to competitive results when the ratio of missing values is limited (typically $<50$\%), but the custom CSDI stands out for the stability of its accuracy when the ratio of missing values increases. It is worth noticing that the original CSDI model leads to disappointing results when applied to this synthetic dataset mimicking high-resolution hydrological observations, most likely because of the limited training sample size, combined with the high dimensionality (75 variables) and the heterogeneous scaling of hydrological signals, as described in section~\ref{subsec_augmentedCSDI}.

Results in Figure~\ref{fig:syn_data_res_graph}.b focus on the relative forecasting performance of the different models across multiple forecast horizons, and show that for this task too the custom CSDI model almost consistently outperforms all the baselines (see Appendix Tables \ref{fig:synthetic_all_results} for more detailed results). For short-term prediction the GP model performs well (second behind custom CSDI) but as the horizon grows the probabilistic TS-Mixer model (TS-Mixer P) gets better and equals the custom CSDI model for the 18 time steps horizon, even marginally surpassing it in terms of RMSE. The good performance of these two models for this prediction horizon could be explained by the fact that the multi-layer perceptron architecture which makes up TS-Mixer is more stable during training than the base CSDI.

\begin{figure}[H]
    \centering
    \includegraphics[width=0.85\linewidth]{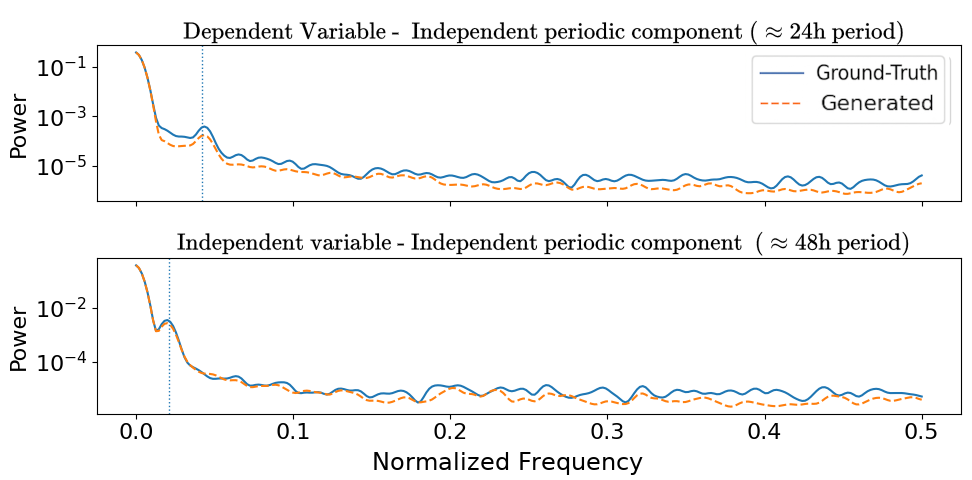}
    \caption{Power spectral density comparison between synthetic data (blue) and the mean of 100 simulations performed by the custom CSDI model (orange). The dotted vertical blue lines show the expected frequencies of the periodic signal embedded in the synthetic data, i.e., 24h for variable 15 and at 48h for variable 18.}
    \label{fig:decomposition_seasons}
\end{figure}

To refine the evaluation of the custom CSDI model, Figure~\ref{fig:decomposition_seasons} investigates how it captures periodic signals through the comparison of the power spectral density (PSD) in simulations and in the synthetic data. For a time series $x$ of length $T$ the PSD is defined as the squared magnitude of the Fourier coefficient, i.e., $PSD(x) = |\sum^T_{t=0} x(t) e^{-j2\pi\frac{t}{T}}|^2$. Spectra are normalized and a one dimensional Gaussian filter kernel is applied along the frequency axis to reduce noise. For the generated time series, the PSD is computed for each individual simulated time series, then the mean PSD over 100 simulations is computed and reported in the figure.

The results in Figure~\ref{fig:decomposition_seasons} show that the custom CSDI model successfully captures and reproduces the daily (dependent periodic variable, top in Figure~\ref{fig:synthetic_data_}) and two-days (independent periodic variable, bottom in Figure~\ref{fig:synthetic_data_}) periodic signals present in the data. Both the amplitude and the frequency are properly modeled. Since the targeted tasks are imputation within short windows (48 timesteps, with 90\% of values missing at random) and short-term forecasting (horizons of at most 18 steps), every evaluation window is shorter than the dominant seasonal period. We therefore do not assess the ability of the custom CSDI model to reproduce low-frequency components such as seasonal or pluriannual cycles.

Finally, Table~\ref{tab:imputation_results} evaluates how imputation accuracy scales when the length of the missing segments (i.e., the size of the data gaps) increases from 50 to 180 consecutive time steps (i.e. from 8.3 days to 30 days). The results reveal a clear performance inflection for the custom CSDI model based on gap size. For short-to-medium horizons (50 and 100 steps), the custom CSDI model outperforms the GP benchmark, achieving a SMAPE of 9.52\% for missing segments of size 100. However, when the gaps exceed 140 steps, the performance of the custom CSDI model sharply decreases with SMAPE rising from 9.52\% (gap size = 100) to 17.78\% (gap size = 140). For such long gaps the GP benchmark achieves better accuracy. These results could be explained by the fact that the custom CSDI model efficiently leverages the local context (i.e., the data available before and after a gap) for the imputation of missing values, but is less effective in capturing the large-scale context. In case of data comparable to sub-daily resolution hydrological time series (i.e., the synthetic dataset presented in this section) the above features make the custom CSDI model well suited for the imputation of gaps of moderate size and for short- to medium-term forecasting (less than two weeks). When local information becomes rare or absent a decrease in performance is expected. 

\begin{table}[htbp]
\centering
\small
\caption{Imputation performance comparison between the proposed Custom Model and Gaussian Processes (GP) across varying missing gap sizes on synthetic time series.}
\label{tab:imputation_results}
\begin{tabular}{lcccc}
\toprule
\textbf{Model / Metric} & \textbf{50 Timesteps} & \textbf{100 Timesteps} & \textbf{140 Timesteps} & \textbf{180 Timesteps} \\
\midrule
\textit{Custom Model} & & & & \\
\quad RMSE & \textbf{0.0405} $\pm$ 0.0006 & \textbf{0.0355} $\pm$ 0.0003 & 0.0718 $\pm$ 0.0006 & 0.0737 $\pm$ 0.0011 \\
\quad MAE  & \textbf{0.0250} $\pm$ 0.0002 & \textbf{0.0241} $\pm$ 0.0001 & 0.0471 $\pm$ 0.0004 & 0.0494 $\pm$ 0.0004 \\
\quad SMAPE (\%) & \textbf{8.3586} $\pm$ 0.0726 & \textbf{9.5221} $\pm$ 0.0387 & 17.7802 $\pm$ 0.1686 & 19.8093 $\pm$ 0.2360 \\
\quad CRPS  & \textbf{0.0555} $\pm$ 0.0009 & \textbf{0.0570} $\pm$ 0.0005 & 0.1173 $\pm$ 0.0009 & 0.1199 $\pm$ 0.0008 \\
\midrule
\textit{Gaussian Processes} & & & & \\
\quad RMSE & 0.0494 $\pm$ 0.0020 & 0.0612 $\pm$ 0.0010 & \textbf{0.0637} $\pm$ 0.0008 & \textbf{0.0660} $\pm$ 0.0010 \\
\quad MAE  & 0.0347 $\pm$ 0.0008 & 0.0447 $\pm$ 0.0007 & \textbf{0.0423} $\pm$ 0.0007 & \textbf{0.0477} $\pm$ 0.0009 \\
\quad SMAPE (\%) & 11.1288 $\pm$ 0.2047 & 13.7765 $\pm$ 0.3650 & \textbf{13.9469} $\pm$ 0.2436 & \textbf{16.7788} $\pm$ 0.4187 \\
\quad CRPS  & 0.0828 $\pm$ 0.0014 & 0.1021 $\pm$ 0.0013 & \textbf{0.1096} $\pm$ 0.0010 & \textbf{0.1216} $\pm$ 0.0016 \\
\bottomrule
\end{tabular}
\end{table}

\subsection{Performance on the OPE dataset}

%\vspace{2mm}

% ===================== FIGURES =====================
%\begin{minipage}{0.49\linewidth}
%\centering
%\includegraphics[width=\linewidth]{figures/ope_imputation_memtrics_graph.png}
%\caption*{(c) Imputation Metrics Visualization}
%\end{minipage}
%\hfill
%\begin{minipage}{0.49\linewidth}
%\centering
%\includegraphics[width=\linewidth]{figures/ope_forecasting_metrics_graph.png}
%\caption*{(d) Forecasting Metrics Visualization}
%\end{minipage}

%\vspace{-3mm}

\begin{figure}[H]
    \centering
    \includegraphics[width=0.5\linewidth]{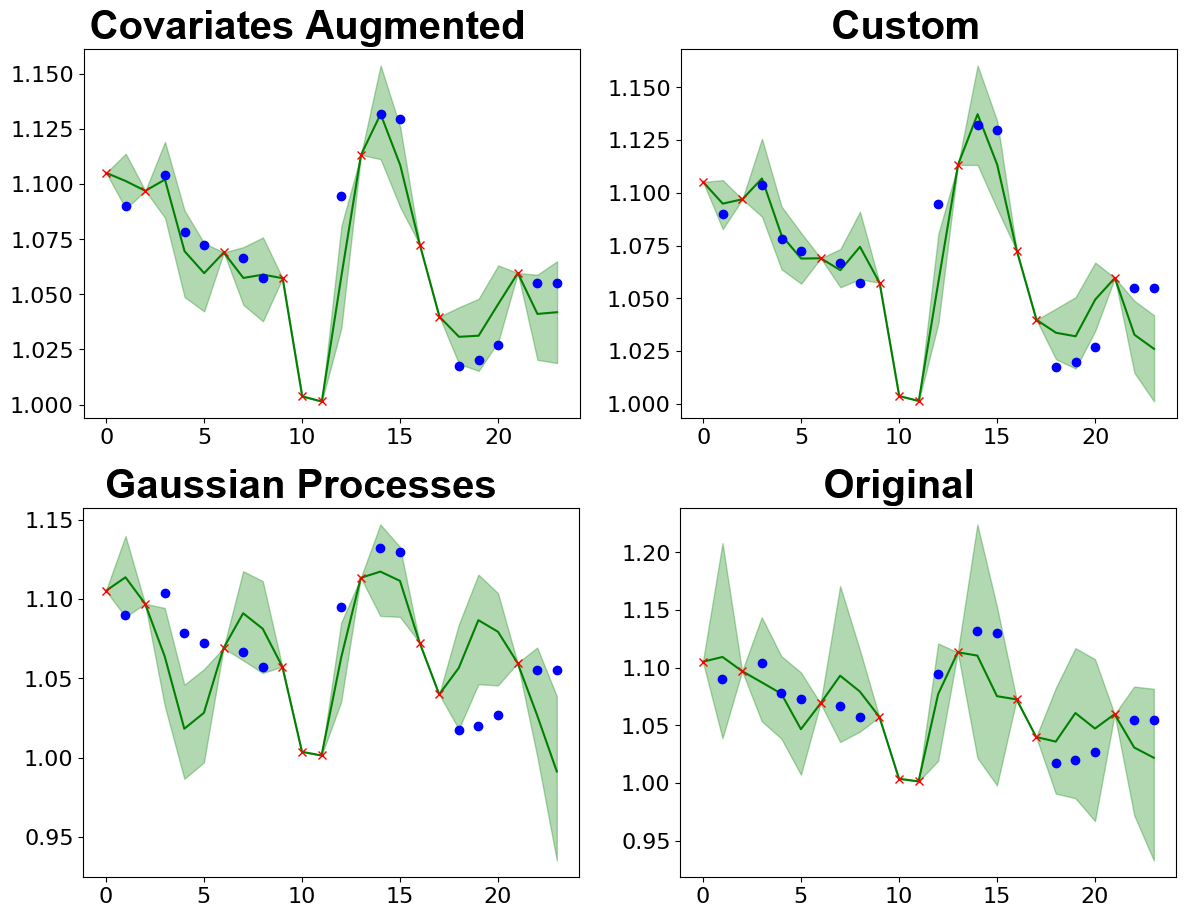}
    \caption{Illustration of imputation results over 24 timesteps (4 days), on one sample of the test dataset, across four models for dissolved oxygen at station OPE90013 showed as normalized values. Red crosses denote observed timesteps, blue dots indicate masked ground truth, and the simulated distribution is represented by its median (green line) and 5-95th percentile band (light green band).}
    \label{fig:imputation_visualization}
\end{figure}

The OPE dataset is partitioned chronologically into training, validation, and test sets of respectively 4152, 20, and 28 elementary time series, each comprising 24 timesteps (4 days). Figure \ref{fig:imputation_visualization} illustrates imputation results for four different models: the original CSDI model, the custom CSDI model, the custom CSDI model augmented with covariates, and the GP benchmark model.

The visual inspection of Figure~\ref{fig:imputation_visualization} indicates that both versions of the custom CSDI model, and to a lesser extent the GP model,  successfully capture the main temporal patterns of the time series. In contrast the original CSDI model tends to overestimate the temporal variability of the underlying process. 

\begin{figure}[H]
% ===================== FIGURES =====================
\begin{minipage}{0.5\linewidth}
\centering
\includegraphics[width=\linewidth]{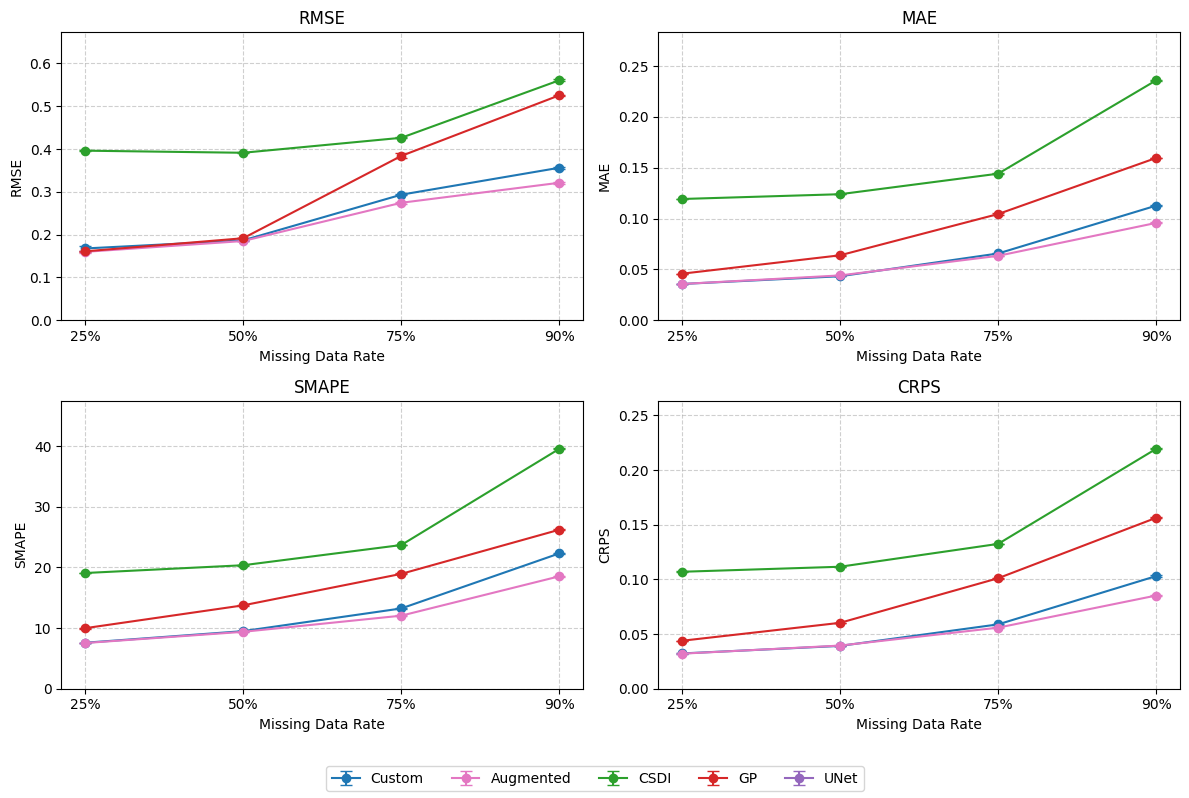}
\caption*{(a) Imputation Metrics Visualization}
\end{minipage}
\hfill
\begin{minipage}{0.5\linewidth}
\centering
\includegraphics[width=\linewidth]{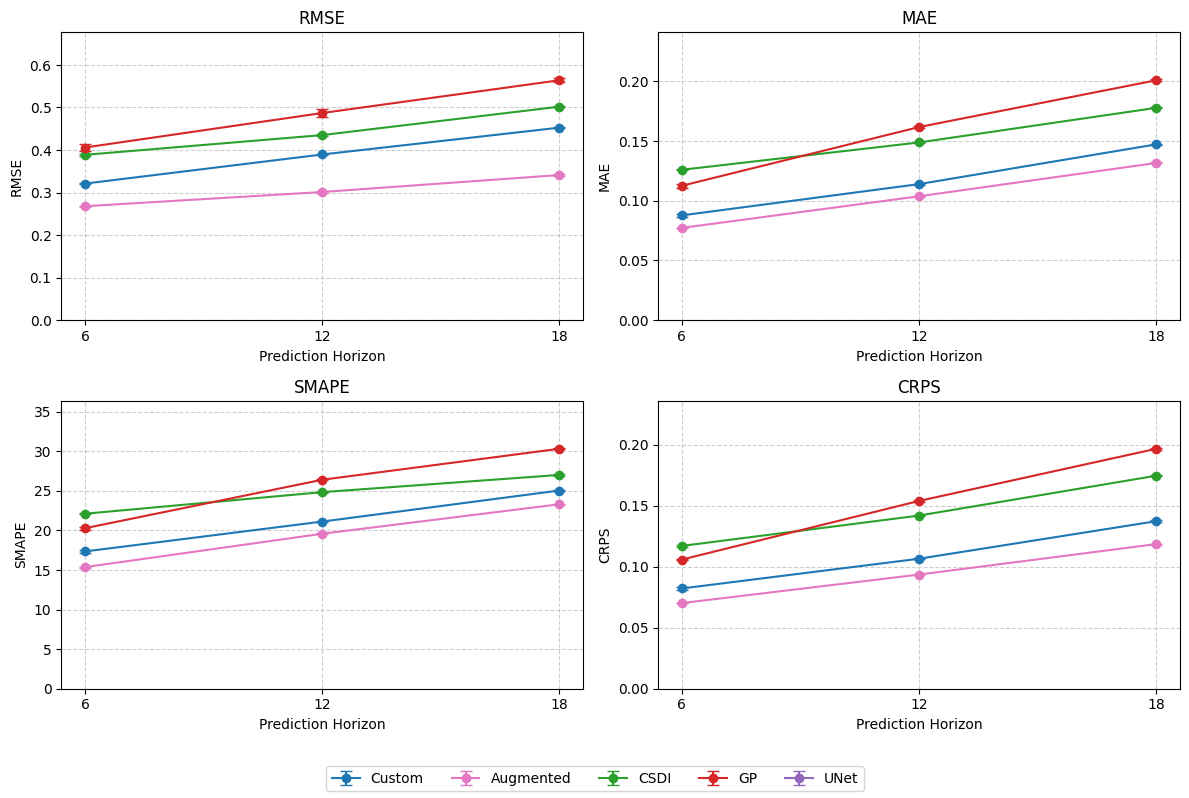}
\caption*{(b) Forecasting Metrics Visualization}
\end{minipage}
\caption{
Performance comparison across missing ratios and prediction horizons on the OPE dataset. U-Net is omitted, as its poor performance would distort the scale of the plots.
}
\label{fig:ope_data_res_graph}
\end{figure}

Figure \ref{fig:ope_data_res_graph} synthesizes the evaluation metrics for both imputation and forecasting, and  detailed results are reported in Appendix Tables \ref{fig:all_results_real}. 

Results in Figure~\ref{fig:ope_data_res_graph} show that across all metrics, the custom CSDI architecture (with and without covariates) consistently outperforms all baselines. As expected, the covariate-augmented version performs slightly better than the variant without covariates, highlighting the additional predictive value of weather auxiliary data. Interestingly, as was the case for the synthetic dataset, the GP model outperforms the original CSDI architecture. This suggests that the original CSDI model may not be suited for hydrological data. In contrast, a custom architecture specifically designed and tailored to the characteristics of the dataset demonstrates superior performance. 

\begin{figure}[H]
    \centering
    \includegraphics[width=0.65\linewidth]{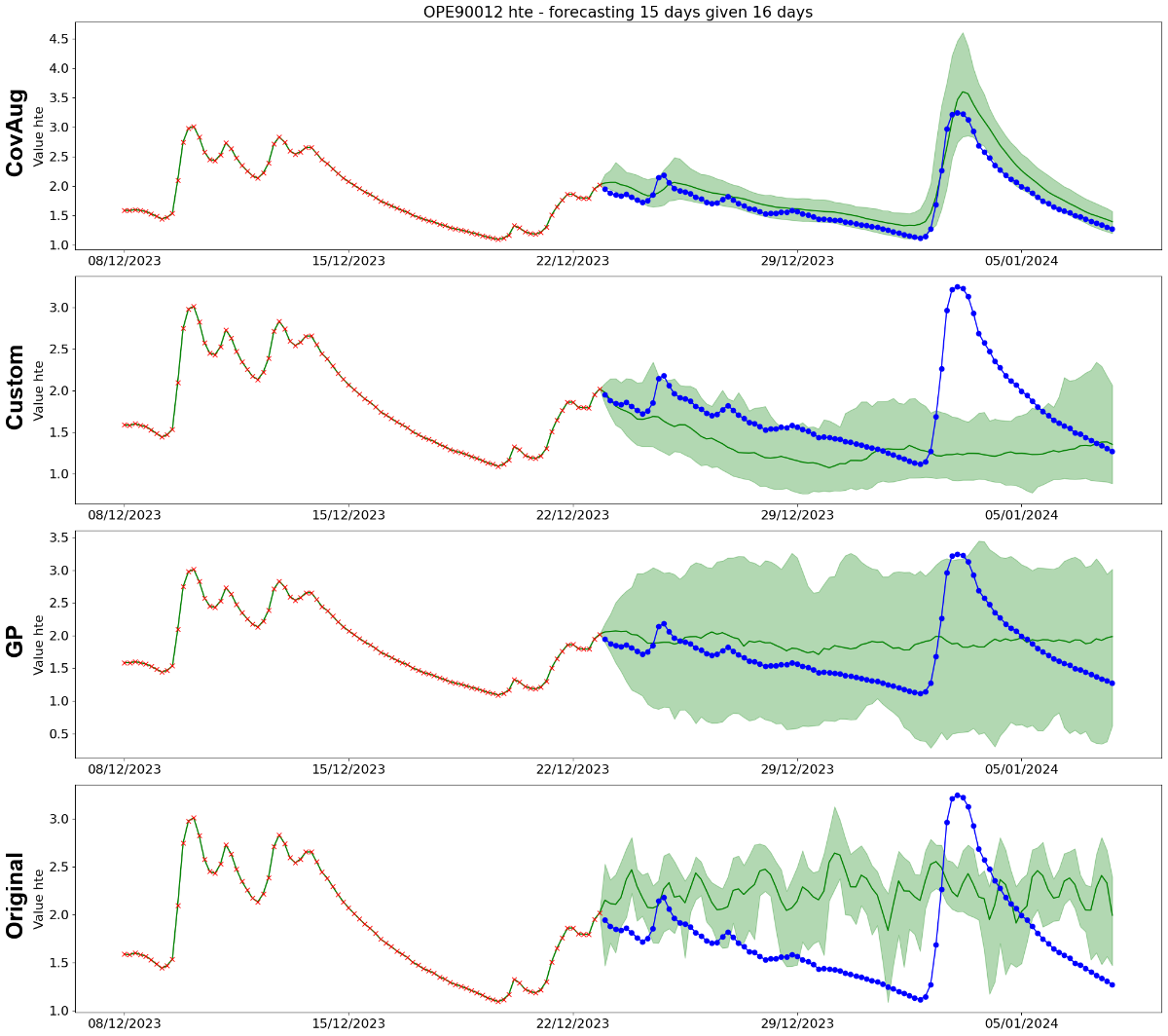}
    \caption{Compared forecasting results for four tested models, Custom with covariates (CovAug), Custom without covariates (Custom), Gaussian Processes (GP) and Original CSDI (Original). Red crosses show observed timesteps, blue dots indicate masked ground truth, and the simulated distribution is represented by its median and 5-95th percentile band. Results are shown for the normalized water level variable (HTE) at station OPE90012.}
    \label{fig:long_term_forecasting}
\end{figure}

The covariate-augmented model performs similarly to the model without covariates in terms of imputation; however, it performs significantly better for forecasting tasks. This improvement is particularly visible for the simulations displayed in Figure \ref{fig:long_term_forecasting}. In this configuration a sudden meteorological event (here a rainfall event) breaks the stationarity of the hydrometeorological system, and the baseline model fails to capture the sudden peak of stream water level observed on 05/01/2024 whereas the model incorporating covariates successfully reproduces this feature thanks to the auxiliary information brought by the meteorological covariates.

\begin{figure}[H]
    \centering
    \includegraphics[width=0.8\linewidth]{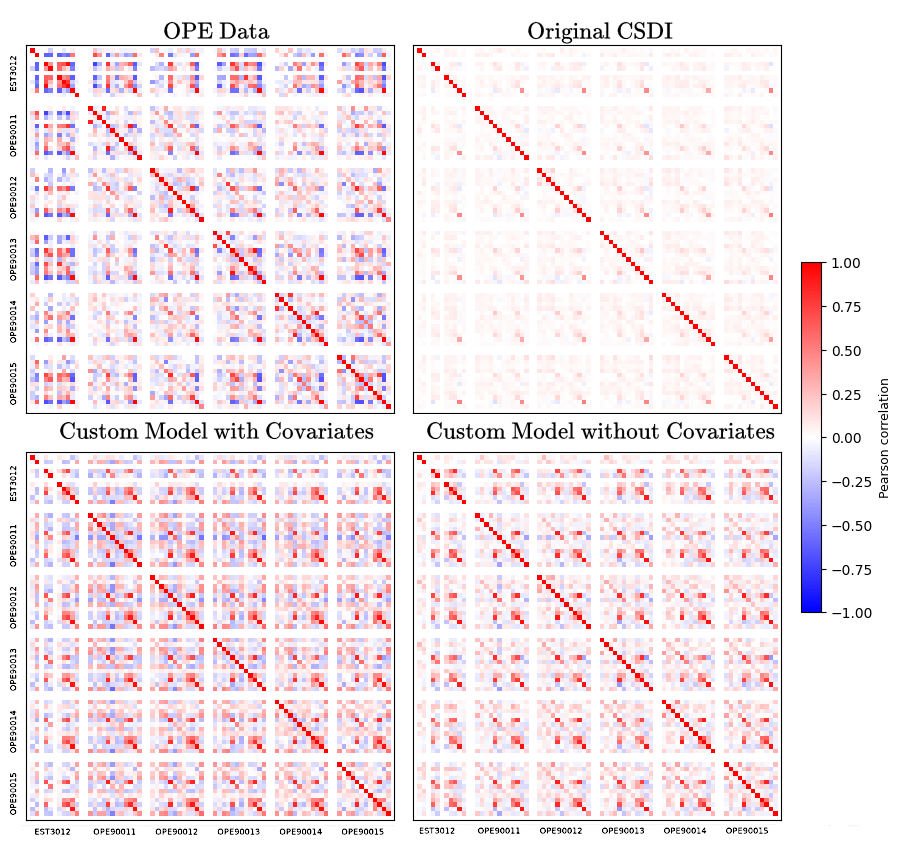}
    \caption{Comparison between the Pearson covariance matrix of the training dataset (top-left) and generated simulations by: the original CSDI (top-right), the custom covariate augmented model (bottom-left), and the custom model (bottom-right). The correlations are grouped by station.}
    \label{fig:correlation_matrix}
\end{figure}

To evaluate the consistency of multivariate time series simulation the Figure~\ref{fig:correlation_matrix} displays the covariance structure of the 75 variables, grouped by station. Parameters are grouped into blocks to facilitate the visualization but this display does not reflect any block-wise modeling assumptions. 

Results in Figure~\ref{fig:correlation_matrix} show that the original CSDI (top-right) struggles to capture the correlation structure. Some emerging structure is visible, with parallel positive diagonals slightly appearing, but overall the simulated correlation seems weaker than the one from the real data. In contrast, the custom model (bottom-right) better reproduces the overall correlation structure across stations. The presence of parallel positive diagonals inside the non-diagonal blocks shows correlations between the same parameters measured at different locations, which suggests that the custom CSDI model performs satisfactorily for multisite modeling. However, the contrast between station is reduced, with weaker correlation magnitudes in custom CSDI simulations compared to observations, suggesting that the generative model partially underestimates the strong positive and negative correlations. In particular, we note that the negative correlations are not well captured. The possible cause for this has already been identified in the literature, as \cite{qin2022cosformerrethinkingsoftmaxattention} showed that since softmax attention weights are strictly non-negative, the cross-station conditioning mechanism struggles to encode negative relationships between variables. Finally, the covariate-augmented CSDI model (bottom-left), which has the same backbone as the custom CSDI model, seems to capture the correlations more strongly. It also captures the parallel positive diagonals inside the blocks, showing correlations between the same parameters measured at different locations. However, in this case, the simulated correlation seems slightly stronger than in the real data.

\begin{figure}[H]
    \centering
    \includegraphics[width=0.95\linewidth]{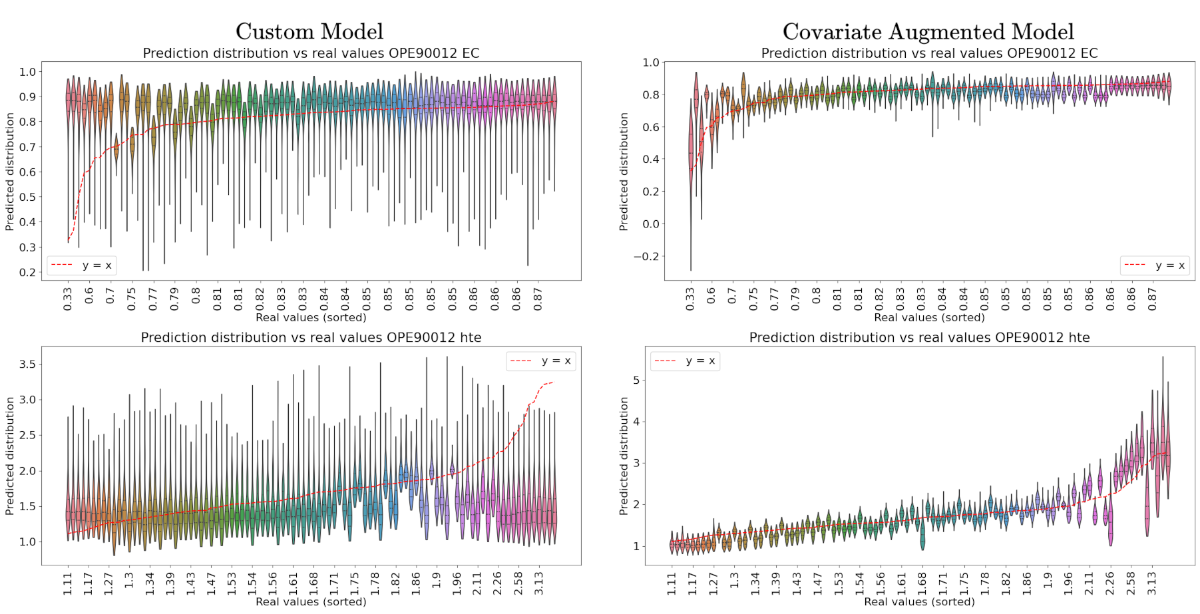}
    \caption{Visualization of the distribution of predicted values with respect to their real value, without covariates (left) and with covariates (right). All values are normalized. The red dashed line represents a perfect predictor. EC stands for electric conductivity, and HTE for water level.}
    \label{fig:distribution_coverage}
\end{figure}

To further explore the contribution of adding covariates to the custom CSDI model, Figure~\ref{fig:distribution_coverage} displays the distribution of simulated values with respect to the observed ones on a forecasting scenario. The red dashed line represents a perfect predictor and is shown for reference.

%Results in Figure~\ref{fig:distribution_coverage} show that the model without covariates has a greater variability, leading to a lower predictive accuracy. This behavior is expected, since the model relies on less information. However, the larger spread also provides broader coverage of possible outcomes, which is desirable in applications where uncertainty quantification is important. In contrast, the model with covariates demonstrates the opposite tradeoff. Predictions are more concentrated around the true values, leading to improved predictive accuracy, but with reduced variability.

To evaluate the ability of the model to generate realistic predictive distributions, this experiment considers the same forecasting task as Figure \ref{fig:long_term_forecasting}, simulating 15 days given 16 days. Results in Figure~\ref{fig:distribution_coverage} highlight the impact of incorporating covariates on the forecasting performance. Without covariates (left), the model has access only to the historical distribution of the target variables and therefore has limited ability to predict the true values. As a consequence, the generated predictions show larger variability and weaker alignment with the ground truth.  However, the larger spread also provides broader coverage of possible outcomes, which is desirable in applications where uncertainty quantification is important. In contrast, the model with covariates demonstrates the opposite tradeoff. Predictions are more concentrated around the true values, leading to improved predictive accuracy, but with reduced variability.

\section{Discussion}
In this study we proposed a Transformer-based diffusion model for the joint imputation and forecasting of multivariate hydrological time series. The resulting model architecture, showcased in \ref{sec:custom_model}, is based on the addition of layers from recent literature to the original CSDI framework in order to tailor the model to the targeted application. We discuss below the strengths and limitations of the custom CSDI model.\\

\textbf{Uncertainty quantification.} Using a diffusion generative model presents the advantage of probabilistic formulation. The model can generate a large ensemble of equally likely simulations, allowing users to quantify uncertainty. This feature is important in the context of hydrology where decision-making is better informed when information about model uncertainty is available. This also allows the model to be used for anomaly detection when values are measured outside the range of simulated values.\\

\textbf{Domain specific architecture.} Through the comparison with the original CSDI and the U-NET architecture, we show that the proposed ad-hoc architecture performs better on both intended tasks, namely imputation and forecasting. Good performance has been reached by drawing on the deep learning literature applied to hydrological time series to guide design choices, such as periodic encoding. At the same time it is important not to lose sight of the conventional generative modeling in order to benefit from research in this community.\\

\textbf{Modeling inter-variable dependencies.} One limitation identified during the evaluation of the custom CSDI model is its limited ability to capture negative cross-variable correlations. While the Transformer architecture has demonstrated strong performance in modeling long-range dependencies through self-attention, the results suggest that the implementation we used has limited capability to capture negative relationships. An interesting direction to address this would be to replace the standard attention mechanism with cosFormer, which has been shown to better handle negative correlations \citep{qin2022cosformerrethinkingsoftmaxattention}. Due to time constraints, this direction was not explored in the current study and is left for future work\\

\textbf{Choice of covariates.} The covariate-augmented model relies on meteorological information from the SAFRAN reanalysis. While this provides a useful way to assess the potential contribution of exogenous meteorological drivers, it does not fully correspond to an operational forecasting setting where future covariates would have to be provided by numerical weather forecasts and would therefore be uncertain. The operational performance of the model may thus depend on the quality of the available meteorological forecasts. This should be evaluated explicitly before using the covariate-augmented custom CSDI model for operational hydrometeorological forecasting.

\textbf{Data Quality.} We note that the OPE dataset used for training was corrected for sensor drift and malfunction through a joint LNE/Andra metrology effort, involving monthly quality control and regular maintenance over 15 years \citep{Guigues_inpress}. This correction is non-trivial: in natural streams, long-term sensor deployment is susceptible to drift from biofouling, sediment, and vegetation debris accumulating on the sensor surface, which can substantially degrade data quality without rigorous, expert-driven QC. While real-time forecasting with the trained model remains possible using raw, uncorrected sensor data, we expect simulation quality to degrade to the extent that inputs deviate from the drift-corrected training distribution.

\section{Conclusion}
In this work, we introduced a Transformer-based architecture able to perform simultaneously the imputation and forecasting of multivariate hydrological time series. The model is based on the CSDI architecture, which has been improved with convolutions, multi-scale layers, and Root Mean Squared Normalization. The resulting architecture allows for a stable training and performs well in modeling synthetic and real datasets. In particular, the custom CSDI model better captures uncertainty than a set of baselines, and is competitive on predictive metrics.

Despite an imperfect modeling of the correlation between variables, the proposed approach offers a promising direction for data-driven modeling of hydrological data with sub-daily resolution across more than 15 years. By jointly addressing imputation and forecasting within a single Transformer-based model using a long-term water quality and water level dataset, this work contributes to the growing literature on deep learning hydrology.

%A possible improvement to our model is to use cosFormer to improve negative correlations capture. In the future, it will be important to improve the model's understanding of the data, for example, by including spatial information in the model.

%A limitation of our study is the use of reanalysis data for our covariates. This is not representative of a real setting, where a predictive model would be used to generate unknown covariates. This means that, in reality, less reliable information will be available to the model, but we believe that this would not change the predictive quality of the model. Still, our study underscores the potential of Transformers for hydrological modeling.

The results support the relevance of the proposed model for short-term forecasting and for the imputation of gaps of size between hours to days. The results support the relevance of the proposed model for short-term forecasting and for the imputation of short to medium-size gaps, but we did not tested the ability of the model to reproduce the hydrological dependencies, seasonal variability, or low-frequency dynamics for periods longer than 15 days. Future work should therefore focus on the evaluation of the model on longer sequences, including long consecutive gaps and temporally structured sensor failures.

% An improvement of the prediction accuracy
% The predictive model using
% Higher than average
% The elements that improve the performances are
% We confirm this hypothesis

% Safran data doesn't represent reality where we would replace w  ith a weather prediction model instead of observed data, giving an edge to our model

% Opens way for further research

\section{Acknowledgment}
This method has been developed within the scope of the Geolearning chair. The authors are grateful to French National Agency for Radioactive Waste Management (ANDRA) partners for support and fruitful discussions. We would especially like to thank their technicians, engineers and researchers responsible for the collection, data curation and quality control of the OPE data, following during more than 15 years a protocol developed in partnership with LNE and ANDRA, with monthly quality control, sensor drift correction, and uncertainty estimation.

\section{Code availability}
The source code is available on this git repository: \href{https://github.com/Pumafi/transformer_diffusion_hydrology}{link}.

\newpage

\bibliographystyle{chicago}
\bibliography{bibliography.bib}

\newpage
\appendix
\section{Gaussian Processes details}
\label{app:gp}

For our baseline, we use a Gaussian Process with a composite kernel combining a smooth periodic component, a non-periodic short-timescale component, and observational noise. Specifically, the kernel consists of a Mat\'ern-$\nu=3/2$ covariance modulated by a periodic exponential-sine-squared term to capture quasi-periodic structure, plus an additional Mat\'ern component for medium-scale variability, and a white-noise term for uncorrelated measurement errors.

\section{Architecture detail}\label{sec:arch_comparison}
\begin{figure}[ht]
    \centering
    \begin{subfigure}{\textwidth}
        \centering
        \includegraphics[width=0.9\linewidth]{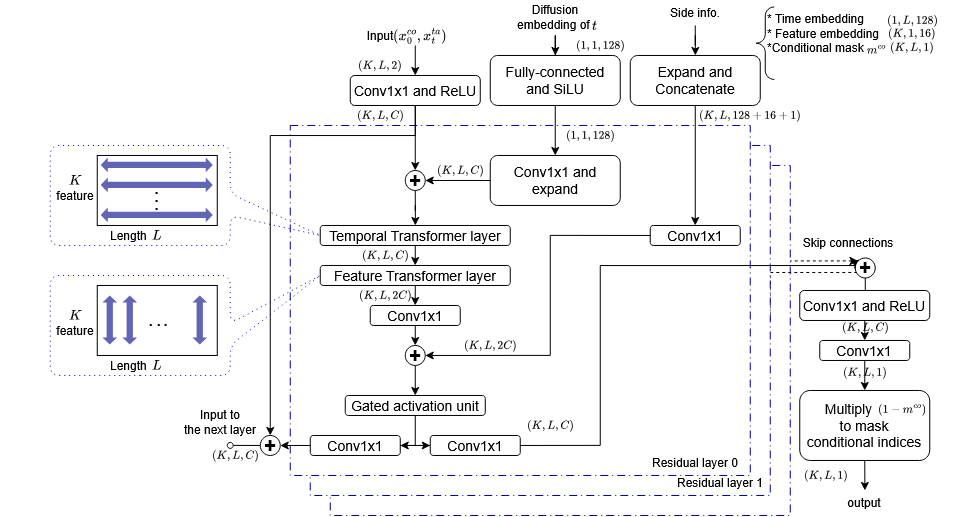}
        \caption{Original architecture}
        \label{fig:arch1}
    \end{subfigure}
    \hfill
    \begin{subfigure}{0.7\textwidth}
        \centering
        \includegraphics[width=\linewidth]{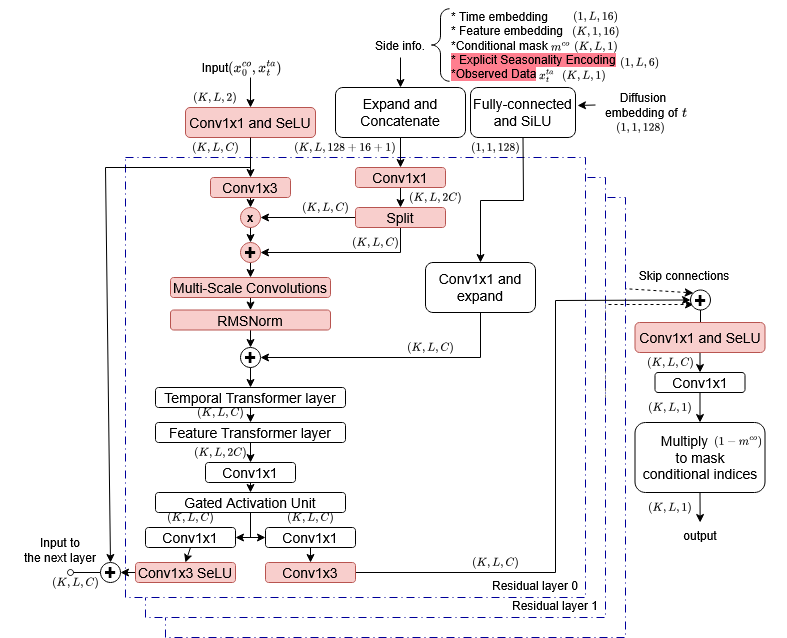}
        \caption{Modified architecture}
        \label{fig:arch2}
    \end{subfigure}
    \caption{Comparison between the original CSDI architecture and the modified architecture. The modified components are highlighted in red. The original architecture is adapted from \cite{CSDI_NEURIPS2021}.}
    \label{fig:arch_comparison}
\end{figure}

\section{Hyperparameters}

\begin{table}[H]
    \centering
    \begin{tabular}{|c|c|c|c|}
        \hline
         Model & Custom & Cov. Augmented & Original \\
         \hline
         Iterations & $100$ & $100$ & $100$ \\
         Recurrent Blocs & $4$ & $4$ & $4$ \\
         Batch Size & $16$ & $16$ & $8$ \\
         Learning Rate & $1e^{-3}$ & $1e^{-3}$ & $1e^{-3}$ \\
         Nb diffusion steps & 100 & 100 & 50 \\
         \hline
    \end{tabular}
    \caption{Hyperparameters used to train the different models.}
    \label{tab:placeholder}
\end{table}

\section{Ablations Tests}\label{sec:ablations_tests}

\begin{table}[H]
\centering
\small
\setlength{\tabcolsep}{3pt}
\begin{tabular}{|l | c | c|c|c|c|c|}
\toprule
\textbf{Model} & \textbf{H}
  & \textbf{RMSE}
  & \textbf{MAE}
  & \textbf{sMAPE}
  & \textbf{CRPS}
  & \textbf{CRPS\_sum} \\
\midrule
% ---- Original ----
\multirow{3}{*}{Original}
  & 6  & $0.3887 \pm 0.0024$ & $0.1260 \pm 0.0003$ & $22.1140 \pm 0.0881$ & $0.1171 \pm 0.0005$ & $0.0656 \pm 0.0001$ \\
  & 12 & $0.4351 \pm 0.0023$ & $0.1489 \pm 0.0005$ & $24.8425 \pm 0.0694$ & $0.1420 \pm 0.0004$ & $0.1020 \pm 0.0001$ \\
  & 18 & $0.5019 \pm 0.0007$ & $0.1779 \pm 0.0004$ & $27.0204 \pm 0.0746$ & $0.1747 \pm 0.0003$ & $0.1528 \pm 0.0002$ \\
\hline
% ---- Cond before Transformers ----
\multirow{3}{*}{\parbox{2.4cm}{\centering +Cond.\ before\\Transformers}}
  & 6  & $0.3257 \pm 0.0008$ & $0.0855 \pm 0.0003$ & $17.5146 \pm 0.1022$ & $0.0802 \pm 0.0004$ & $0.0555 \pm 0.0001$ \\
  & 12 & $0.4009 \pm 0.0005$ & $0.1234 \pm 0.0006$ & $22.8549 \pm 0.1559$ & $0.1159 \pm 0.0008$ & $0.0887 \pm 0.0005$ \\
  & 18 & $0.4733 \pm 0.0005$ & $0.1661 \pm 0.0007$ & $27.7593 \pm 0.1747$ & $0.1558 \pm 0.0005$ & $0.1405 \pm 0.0005$ \\
\hline
% ---- StyleGAN Cond + Norm ----
\multirow{3}{*}{\parbox{2.4cm}{\centering +StyleGAN\\Cond+Norm}}
  & 6  & $0.3240 \pm 0.0007$ & $0.0840 \pm 0.0004$ & \cellcolor{green!25}$\mathbf{16.6410 \pm 0.1247}$ & \cellcolor{green!25}$\mathbf{0.0785 \pm 0.0005}$ & \cellcolor{green!25}$\mathbf{0.0548 \pm 0.0001}$ \\
  & 12 & $0.3929 \pm 0.0008$ & $0.1163 \pm 0.0004$ & $21.5238 \pm 0.0816$ & $0.1087 \pm 0.0004$ & $0.0844 \pm 0.0003$ \\
  & 18 & $0.4610 \pm 0.0011$ & $0.1540 \pm 0.0010$ & $25.6988 \pm 0.0860$ & $0.1429 \pm 0.0009$ & $0.1288 \pm 0.0009$ \\
\hline
% ---- + Conv ----
\multirow{3}{*}{\centering$+$Convolutions}
  & 6  & \cellcolor{green!25}$\mathbf{0.3146 \pm 0.0008}$ & \cellcolor{green!25}$\mathbf{0.0839 \pm 0.0003}$ & $17.7016 \pm 0.0865$ & $0.0794 \pm 0.0002$ & $0.0555 \pm 0.0001$ \\
  & 12 & $0.4018 \pm 0.0009$ & $0.1208 \pm 0.0005$ & $22.0666 \pm 0.1432$ & $0.1137 \pm 0.0005$ & $0.0865 \pm 0.0003$ \\
  & 18 & $0.4830 \pm 0.0004$ & $0.1629 \pm 0.0004$ & $27.2257 \pm 0.0340$ & $0.1544 \pm 0.0004$ & $0.1354 \pm 0.0003$ \\
\hline
% ---- + Seasonality ----
\multirow{3}{*}{\centering$+$Seasonality}
  & 6  & $0.3211 \pm 0.0021$ & $0.0878 \pm 0.0010$ & $17.3480 \pm 0.1479$ & $0.0822 \pm 0.0009$ & $0.0568 \pm 0.0002$ \\
  & 12 & \cellcolor{green!25}$\mathbf{0.3895 \pm 0.0007}$ & \cellcolor{green!25}$\mathbf{0.1140 \pm 0.0004}$ & \cellcolor{green!25}$\mathbf{21.1204 \pm 0.1045}$ & \cellcolor{green!25}$\mathbf{0.1066 \pm 0.0006}$ & \cellcolor{green!25}$\mathbf{0.0846 \pm 0.0004}$ \\
  & 18 & \cellcolor{green!25}$\mathbf{0.4526 \pm 0.0009}$ & \cellcolor{green!25}$\mathbf{0.1473 \pm 0.0004}$ & \cellcolor{green!25}$\mathbf{25.0509 \pm 0.0569}$ & \cellcolor{green!25}$\mathbf{0.1374 \pm 0.0007}$ & \cellcolor{green!25}$\mathbf{0.1272 \pm 0.0006}$ \\
\bottomrule
\end{tabular}
\caption{Ablation study results (mean $\pm$ std) across forecasting horizons.
Bold green indicates the best (lowest) value per horizon/metric pair.}
\label{tab:ablation}
\end{table}

\begin{figure}[H]
    \centering
    \includegraphics[width=0.75\linewidth]{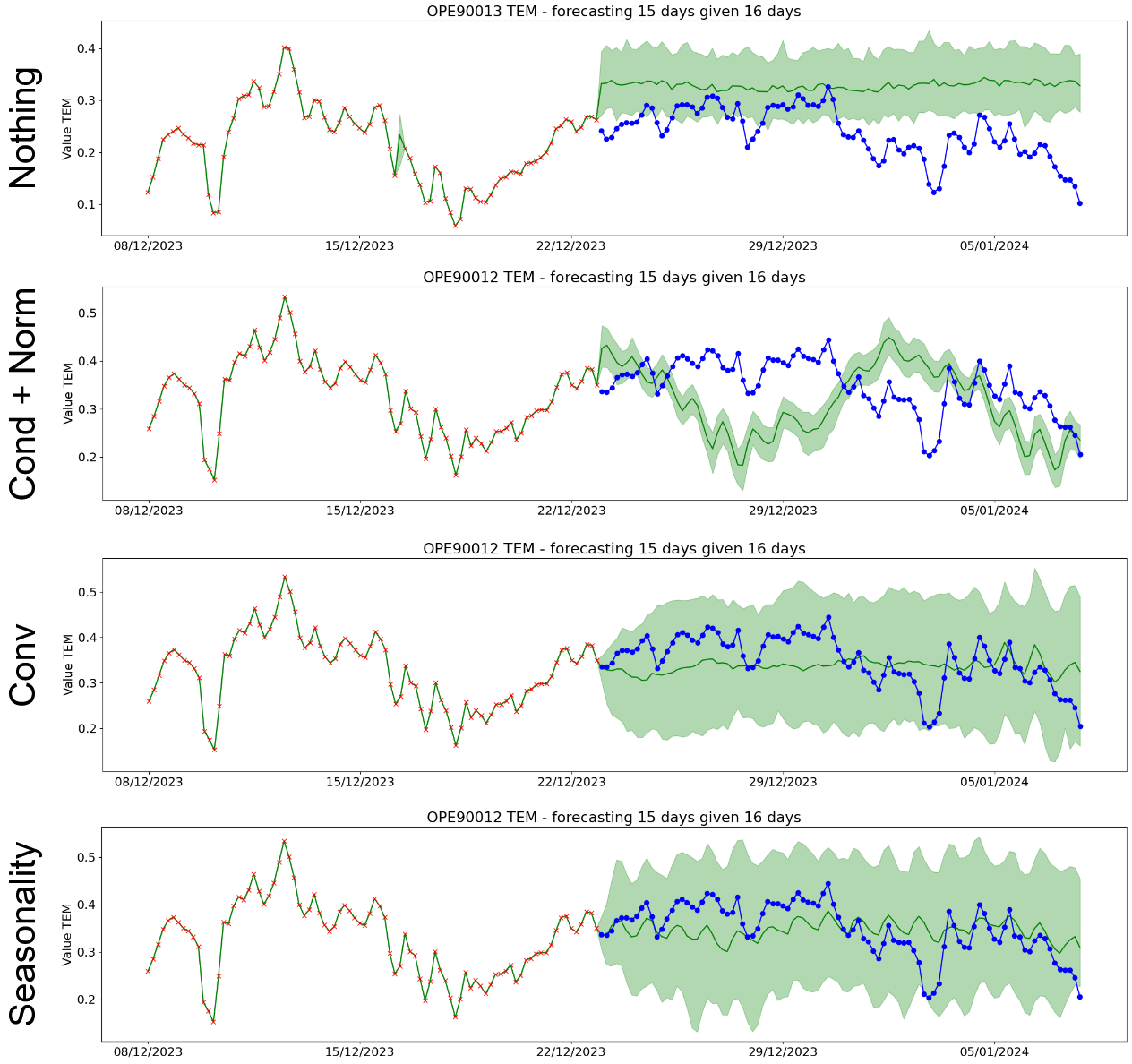}
    \caption{Ablation study: predicted water temperature forecasts (15-day horizon, 
    conditioned on 16 observed days) at station OPE0013, for each 
    successive architectural modification. Green line and shaded envelope: 
    posterior mean and uncertainty band; blue line: ground truth; red crosses: 
    observed conditioning values. Each panel isolates the effect of one added 
    component, from vanilla CSDI (top) to our full proposed architecture 
    (bottom).}
    \label{figure:ablations_tests_vizu}
\end{figure}

In this section, we present some ablation tests to showcase how different layers influence simulation results. Table \ref{tab:ablation} presents forecasting metrics as we add our custom layers, step by step. Figure \ref{figure:ablations_tests_vizu} illustrates forecasting results for these different results.

As discussed in Section~\ref{subsec_originalCSDI}, the vanilla CSDI baseline fails to converge to a meaningful solution: the predicted mean is flat, no temporal structure is captured, and all reported metrics show poor performances.\\

\textbf{Conditioning placement}. Moving the conditioning layer \emph{before} the Transformer blocks, rather than after like in the original CSDI, is the first modification necessary for convergence. This single change enables the model to produce temporally coherent outputs, as seen in Figure \ref{figure:ablations_tests_vizu}, suggesting that early injection of the conditioning signal is critical for the diffusion process to learn meaningful dynamics.\\

\textbf{RMSNorm and StyleGAN-based conditioning}. To improve training stability, we introduce RMSNormalization alongside StyleGAN-style conditioning. Without normalization, gradients explode during training, preventing convergence. We noticed \emph{a posteriori} that this design is closely related to Adaptive Layer Normalization (AdaLN), a technique widely adopted in Diffusion Transformer (DiT) architectures~\citep{peebles2023scalablediffusionmodelstransformers} precisely because it stabilizes transformer training in the diffusion setting. Our variant differs in using RMSNorm rather than standard Layer Normalization and omits the learned scaling factor. We are not the only ones to choose  RMSNorm, as \cite{jones2026elucidatingdesignspaceflow} independently explored it for diffusion transformers. At this stage, the generated time series exhibit realistic hydrological structure, however predictions are substantially biased and the predictive uncertainty does not cover the ground truth.\\

\begin{figure}[H]
    \centering
    \includegraphics[width=0.75\linewidth]{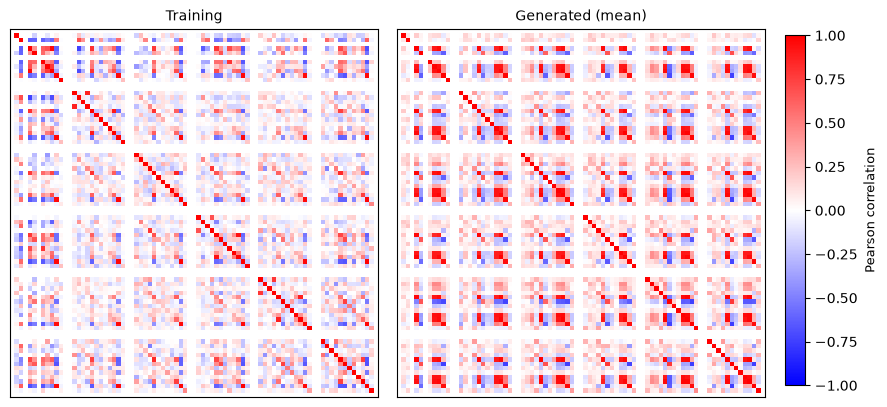}
    \caption{Inter-variable correlation matrices estimated, real-data (left) and our custom architecture without convolutions (right).}
    \label{figure:ablations_tests_correl}
\end{figure}

\textbf{Convolutional layers}. Adding convolutional layers on top of the above yields to a trade-off: the predicted mean becomes slightly flatter, while the confidence band widens considerably, driving the model toward a more uncertain but better-calibrated regime. Qualitatively, the simulated time series still behave like real hydrological time series, yet seasonality remains poorly captured and temporal dependencies appear chaotic (Figure~\ref{figure:ablations_tests_vizu}). Point-wise metrics (MAE, RMSE) improve at short horizons, but all other losses worsen. Critically, however, removing the convolutional layers degrades the inter-variable correlation structure (Figure~\ref{figure:ablations_tests_correl}), indicating that without them the model overfits and fails to reproduce important multivariate dependencies.\\

\textbf{Encoding periodic signals}. Finally, incorporating the encoding of periodic signals yields the best performance on long-horizon forecasting across all metrics. As shown in Figure~\ref{figure:ablations_tests_vizu}, it is only with this component that the predicted peaks align with those of the ground truth, while maintaining a well-calibrated uncertainty range. This constitutes our final custom architecture.

\section{Synthetic dataset details}\label{annex:synthetic}
The synthetic dataset is designed to mimic multivariate hydrological time series at $K$ different stations. It contains three groups of variables per station: a dynamic component, a correlated periodic component, and an independent seasonal component. In addition, a regime-switching process mimics a station with drought behavior. For each station $1\leq k \leq K$, we define a baseline mean $\mu_k = 0.25 + 0.07 k$, modeling differences between measurement locations. The latent driver of each dynamic variable follows an autoregressive AR(1) process,

\begin{equation}
    z_{k, t} = \max\big(0.2, \mu_k + \phi_k (z_{k, t-1} - \mu_k) + \epsilon_{k, t} + g_t\big),  \quad \epsilon_{k, t} \sim \mathcal{N}(0, 0.01^2),
\end{equation}

with AR coefficient $\phi_k = 0.95 - 0.02 k$, and $g_t$ denotes the coefficient for global extremes shared between all stations

\begin{equation}
g_t = 
\begin{cases} 
\text{Exp}(0.2) & \text{with probability } p_{\text{spike}}(t), \\
0 & \text{otherwise},
\end{cases}
\end{equation}

where the probability of spike $p_{\text{spike}}(t)$ is higher during periods of bursts $b_t$ and otherwise low. Bursts evolve according to
\begin{equation}
b_t = 
\begin{cases}
1 & \text{with probability } 0.01, \\
b_{t-1} \cdot \beta & \text{otherwise}.
\end{cases}
\end{equation}

where $\beta = 0.93$ represents the burst decay. Then we define $p_{\text{spike}}(t)$ as:

\begin{equation}
    p_{\text{spike}}(t) = 
\begin{cases}
0.08 & \text{if } b_t > 0.2, \\
0.01 & \text{otherwise}.
\end{cases}
\end{equation}

First, the observed dynamic variable is obtained by adding small observation noise to the latent,
\begin{equation}
D_{k, t} = z_{k, t} + \eta_t, 
\quad \eta_t \sim \mathcal{N}(0, 0.01^2),
\end{equation}

To mimic periodic cycles of hydrological variable, a correlated periodic counterpart to the observed dynamic variable is defined as
\begin{equation}
S_{k, t} = 0.6\, z_{k, t} + s_{k, t} + \xi_t, 
\quad \xi_t \sim \mathcal{N}(0, 0.01^2).
\end{equation}

with $s_{k, t}$ a periodic component composed of two sinusoidal terms with slightly varying periods and random phase shifts $\phi_{1,k}, \phi_{2,k} \sim \text{Uniform}(0, 2\pi)$, defined as

\begin{equation}
s_{k, t} = 0.07 \sin\left(\frac{2 \pi t}{p_{1,k}} + \phi_{1,k} \right) 
+ 0.03 \sin\left(\frac{2 \pi t}{p_{2,k}} + \phi_{2,k} \right).
\end{equation}

Each station also includes a simple independent periodic variable, defined as
\begin{equation}
Z_{k, t} =
0.15
+ 0.08 \sin\left(\frac{2 \pi t}{\tilde p_{1,k}} + \tilde\phi_{1,k} \right)
+ 0.04 \sin\left(\frac{2 \pi t}{\tilde p_{2,k}} + \tilde\phi_{2,k} \right)
+ \omega_t,
\end{equation}

with $\omega_t \sim \mathcal{N}(0, 0.01^2)$. The frequencies and phases of this component are independent of the latent process and global extremes.

In addition to station-level variables, a single regime process $R[t]$ models a station with drought dynamics. In the normal regime, the process follows

\begin{equation}
R_t = \mu_N + \phi_N (R[t-1] - \mu_N) + \epsilon_t + 0.2\, g_t,
\end{equation}

with $\mu_N = 0.35$, $\phi_N = 0.94$, and $\epsilon_t \sim \mathcal{N}(0, 0.01^2)$. During drought periods, the process gradually reverts toward a near-zero mean,

\begin{equation}
R_t = \mu_D + \phi_D (R[t-1] - \mu_D),
\end{equation}

with $\mu_D = 0.02$ and $\phi_D = 0.85$. Drought episodes persist for a minimum duration and evolve smoothly to avoid instantaneous drops.

Spikes and drought are mutually exclusive: if $g_t > 0$, the regime is forced to normal dynamics.

Finally, the data set stacks the three dynamic, three correlated periodic, three independent periodic variables, and the regime process as one multivariate time series.

\begin{equation}
X = 
\begin{bmatrix}
D_1, & \ldots, & D_K, & S_1, & \ldots, & S_K, & Z_1, & \ldots, & Z_K, & R
\end{bmatrix}
\in \mathbb{R}^{n \times (3K+1)},
\end{equation}

\newpage

\section{Detailed metrics}

%%%%%%%%%%%%%%%%%%%%%%%%%%%%%%%%%%%%%%%%%%%%%%%%%%%%%%%%%%%%%%%%%%%%%%%%%%%%%%%%%%%%%%%%%%%%%%%%%
%
% Synthetic
%
%%%%%%%%%%%%%%%%%%%%%%%%%%%%%%%%%%%%%%%%%%%%%%%%%%%%%%%%%%%%%%%%%%%%%%%%%%%%%%%%%%%%%%%%%%%%%%%%%

\begin{figure*}[hb]
\centering
\scriptsize
\setlength{\tabcolsep}{3pt}
\renewcommand{\arraystretch}{0.9}

% ===================== TABLES =====================
\begin{minipage}{0.75\linewidth}
\centering
\resizebox{\linewidth}{!}{
\begin{tabular}{|l|l|c|c|c|c|}
\hline
Missing & Model & RMSE & MAE & SMAPE & CRPS \\
\hline
\multirow{6}{*}{25\%} 
& Custom  & \cellcolor{green!25}0.0119 $\pm$ 0.0000 & \cellcolor{green!25}0.0094 $\pm$ 0.0000 & \cellcolor{green!25}4.2593 $\pm$ 0.0023 & \cellcolor{green!25}0.0208 $\pm$ 0.0000 \\
& GP      & \cellcolor{blue!15}0.0272 $\pm$ 0.0001 & \cellcolor{blue!15}0.0151 $\pm$ 0.0000 & \cellcolor{blue!15}5.4961 $\pm$ 0.0113 & \cellcolor{blue!15}0.0354 $\pm$ 0.0001 \\
& U-NET    & 0.1629 $\pm$ 0.0004 & 0.1372 $\pm$ 0.0004 & 52.4132 $\pm$ 0.2071 & 0.2956 $\pm$ 0.0004 \\
& NHITS   & -- & -- & -- & -- \\
& TSMixer-D/P & -- & -- & -- & -- \\
& CSDI    & 0.0925 $\pm$ 0.0002 & 0.0652 $\pm$ 0.0001 & 21.0319 $\pm$ 0.0289 & 0.1471 $\pm$ 0.0002 \\
\hline
\multirow{6}{*}{50\%} 
& Custom  & \cellcolor{green!25}0.0125 $\pm$ 0.0000 & \cellcolor{green!25}0.0098 $\pm$ 0.0000 & \cellcolor{green!25}4.3723 $\pm$ 0.0031 & \cellcolor{green!25}0.0218 $\pm$ 0.0000 \\
& GP      & \cellcolor{blue!15}0.0323 $\pm$ 0.0001 & \cellcolor{blue!15}0.0175 $\pm$ 0.0000 & \cellcolor{blue!15}6.1521 $\pm$ 0.0150 & \cellcolor{blue!15}0.0411 $\pm$ 0.0001 \\
& U-NET    & 0.2082 $\pm$ 0.0003 & 0.1743 $\pm$ 0.0003 & 70.2633 $\pm$ 0.3969 & 0.3704 $\pm$ 0.0004 \\
& NHITS   & -- & -- & -- & -- \\
& TSMixer-D/P & -- & -- & -- & -- \\
& CSDI    & 0.1269 $\pm$ 0.0004 & 0.0862 $\pm$ 0.0002 & 24.8444 $\pm$ 0.0406 & 0.1987 $\pm$ 0.0006 \\
\hline
\multirow{6}{*}{75\%} 
& Custom  & \cellcolor{green!25}0.0154 $\pm$ 0.0002 & \cellcolor{green!25}0.0110 $\pm$ 0.0000 & \cellcolor{green!25}4.7121 $\pm$ 0.0033 & \cellcolor{green!25}0.0241 $\pm$ 0.0000 \\
& GP      & \cellcolor{blue!15}0.0438 $\pm$ 0.0002 & \cellcolor{blue!15}0.0239 $\pm$ 0.0000 & \cellcolor{blue!15}8.2759 $\pm$ 0.0184 & \cellcolor{blue!15}0.0567 $\pm$ 0.0001 \\
& U-NET    & 0.2342 $\pm$ 0.0001 & 0.1950 $\pm$ 0.0002 & 81.2690 $\pm$ 0.0595 & 0.4147 $\pm$ 0.0004 \\
& NHITS   & -- & -- & -- & -- \\
& TSMixer-D/P & -- & -- & -- & -- \\
& CSDI    & 0.1482 $\pm$ 0.0006 & 0.0998 $\pm$ 0.0004 & 28.4976 $\pm$ 0.0813 & 0.2357 $\pm$ 0.0008 \\
\hline
\multirow{6}{*}{90\%} 
& Custom  & \cellcolor{green!25}0.0237 $\pm$ 0.0002 & \cellcolor{green!25}0.0157 $\pm$ 0.0001 & \cellcolor{green!25}6.7905 $\pm$ 0.0257 & \cellcolor{green!25}0.0339 $\pm$ 0.0001 \\
& GP      & \cellcolor{blue!15}0.0687 $\pm$ 0.0002 & \cellcolor{blue!15}0.0400 $\pm$ 0.0001 & \cellcolor{blue!15}14.3499 $\pm$ 0.0307 & \cellcolor{blue!15}0.0959 $\pm$ 0.0003 \\
& U-NET    & 0.2461 $\pm$ 0.0003 & 0.2048 $\pm$ 0.0002 & 86.5986 $\pm$ 0.1939 & 0.4352 $\pm$ 0.0003 \\
& NHITS   & -- & -- & -- & -- \\
& TSMixer-D/P & -- & -- & -- & -- \\
& CSDI    & 0.2250 $\pm$ 0.0005 & 0.1734 $\pm$ 0.0005 & 48.3544 $\pm$ 0.1071 & 0.3952 $\pm$ 0.0012 \\
\hline
\end{tabular}}
\caption*{(a) Synthetic Imputation Metrics}
\end{minipage}

\hfill

\begin{minipage}{0.75\linewidth}
\centering
\resizebox{\linewidth}{!}{
\begin{tabular}{|l|l|c|c|c|c|}
\hline
Horizon & Model & RMSE & MAE & SMAPE & CRPS \\
\hline
\multirow{7}{*}{6}
& Custom   & \cellcolor{green!25}0.0489 $\pm$ 0.0000 & \cellcolor{green!25}0.0207 $\pm$ 0.0000 & \cellcolor{green!25}6.6687 $\pm$ 0.0121 & \cellcolor{green!25}0.0512 $\pm$ 0.0002 \\
& GP       & \cellcolor{blue!15}0.0529 $\pm$ 0.0007 & \cellcolor{blue!15} 0.0284 $\pm$ 0.0002 & 9.8296 $\pm$ 0.0487 & \cellcolor{blue!15} 0.0699 $\pm$ 0.0005 \\
& U-NET     & 0.2404 $\pm$ 0.0011 & 0.2013 $\pm$ 0.0013 & 85.5644 $\pm$ 0.7897 & 0.4319 $\pm$ 0.0020 \\
& NHITS    & 0.0578 $\pm$ 0.0000 & 0.0367 $\pm$ 0.0000 & 15.2595 $\pm$ 0.0000 & 0.1062 $\pm$ 0.0000 \\
& TSMixer-D& 0.0524 $\pm$ 0.0000 & 0.0274 $\pm$ 0.0000 & 10.3289 $\pm$ 0.0000 & 0.0793 $\pm$ 0.0000 \\
& TSMixer-P& 0.0588 $\pm$ 0.0010 & 0.0295 $\pm$ 0.0004 & \cellcolor{blue!15} 9.7174 $\pm$ 0.0434 & 0.0893 $\pm$ 0.0004 \\
& CSDI     & 0.1889 $\pm$ 0.0005 & 0.1345 $\pm$ 0.0005 & 34.8726 $\pm$ 0.1201 & 0.3154 $\pm$ 0.0011 \\
\hline
\multirow{7}{*}{12}
& Custom   & \cellcolor{green!25}0.0565 $\pm$ 0.0001 & \cellcolor{green!25}0.0263 $\pm$ 0.0001 & \cellcolor{green!25}8.1881 $\pm$ 0.0186 & \cellcolor{green!25}0.0635 $\pm$ 0.0007 \\
& GP       & 0.0770 $\pm$ 0.0004 & 0.0442 $\pm$ 0.0002 & 14.5196 $\pm$ 0.0939 & 0.1047 $\pm$ 0.0004 \\
& U-NET     & 0.2484 $\pm$ 0.0008 & 0.2071 $\pm$ 0.0006 & 88.3506 $\pm$ 0.4370 & 0.4413 $\pm$ 0.0010 \\
& NHITS    & 0.0615 $\pm$ 0.0000 & 0.0385 $\pm$ 0.0000 & 15.8211 $\pm$ 0.0000 & 0.1113 $\pm$ 0.0000 \\
& TSMixer-D& 0.0614 $\pm$ 0.0000 & 0.0346 $\pm$ 0.0000 & 13.3507 $\pm$ 0.0000 & 0.1000 $\pm$ 0.0000 \\
& TSMixer-P& \cellcolor{blue!15}0.0582 $\pm$ 0.0001 & \cellcolor{blue!15}0.0314 $\pm$ 0.0001 & \cellcolor{blue!15}10.8374 $\pm$ 0.0271 & \cellcolor{blue!15}0.0822 $\pm$ 0.0001 \\
& CSDI     & 0.1864 $\pm$ 0.0006 & 0.1332 $\pm$ 0.0003 & 35.3875 $\pm$ 0.0216 & 0.3196 $\pm$ 0.0009 \\
\hline
\multirow{7}{*}{18}
& Custom   & \cellcolor{blue!15}0.0714 $\pm$ 0.0000 & \cellcolor{green!25}0.0330 $\pm$ 0.0001 & \cellcolor{green!25}9.5539 $\pm$ 0.0330 & \cellcolor{green!25}0.0794 $\pm$ 0.0006 \\
& GP       & 0.0823 $\pm$ 0.0002 & 0.0484 $\pm$ 0.0001 & 16.8145 $\pm$ 0.0592 & 0.1159 $\pm$ 0.0003 \\
& U-NET     & 0.2509 $\pm$ 0.0007 & 0.2080 $\pm$ 0.0007 & 88.8613 $\pm$ 0.7188 & 0.4442 $\pm$ 0.0009 \\
& NHITS    & 0.0764 $\pm$ 0.0000 & 0.0456 $\pm$ 0.0000 & 17.2002 $\pm$ 0.0000 & 0.1318 $\pm$ 0.0000 \\
& TSMixer-D& 0.0743 $\pm$ 0.0000 & 0.0430 $\pm$ 0.0000 & 16.0825 $\pm$ 0.0000 & 0.1243 $\pm$ 0.0000 \\
& TSMixer-P& \cellcolor{green!25}0.0682 $\pm$ 0.0002 & \cellcolor{blue!15}0.0361 $\pm$ 0.0001 & \cellcolor{blue!15}11.7899 $\pm$ 0.0247 & \cellcolor{blue!15}0.0865 $\pm$ 0.0001 \\
& CSDI     & 0.1831 $\pm$ 0.0002 & 0.1309 $\pm$ 0.0001 & 34.7984 $\pm$ 0.0380 & 0.3155 $\pm$ 0.0005 \\
\hline
\end{tabular}}
\caption*{(b) Synthetic Forecasting Metrics}
\end{minipage}

\caption{
Synthetic dataset performance comparison. Best values are highlighted in green; second-best in blue.
\textbf{Top}: Imputation metrics across missing ratios.
\textbf{Bottom}: Forecasting metrics across horizons.
}
\label{fig:synthetic_all_results}
\vspace{-4mm}
\end{figure*}

%%%%%%%%%%%%%%%%%%%%%%%%%%%%%%%%%%%%%%%%%%%%%%%%%%%%%%%%%%%%%%%%%%%%%%%%%%%%%%%%%%%%%%%%%%%%%%%%%
%
% OPE
%
%%%%%%%%%%%%%%%%%%%%%%%%%%%%%%%%%%%%%%%%%%%%%%%%%%%%%%%%%%%%%%%%%%%%%%%%%%%%%%%%%%%%%%%%%%%%%%%%%

\begin{figure*}[ht]
\centering
\scriptsize
\setlength{\tabcolsep}{3pt}
\renewcommand{\arraystretch}{0.9}

% ===================== TABLES =====================
\begin{minipage}{0.75\linewidth}
\centering
\resizebox{\linewidth}{!}{
\begin{tabular}{|l|l|c|c|c|c|}
\hline
Missing & Model & RMSE & MAE & SMAPE & CRPS \\
\hline
\multirow{5}{*}{25\%}
& Custom & 0.1676 (±0.0049) & \cellcolor{blue!15}0.0356 (±0.0003) & 7.5701 (±0.0195) & 0.0322 (±0.0003) \\
& Augmented & \cellcolor{green!25}0.1597 (±0.0029) & \cellcolor{blue!15}0.0356 (±0.0002) & \cellcolor{green!25}7.5243 (±0.0260) & \cellcolor{green!25}0.0320 (±0.0002) \\
& GP & \cellcolor{blue!15}0.1609 (±0.0008) & 0.0457 (±0.0002) & \cellcolor{blue!15}9.9667 (±0.0560) & \cellcolor{blue!15}0.0438 (±0.0001) \\
& CSDI & 0.3962 (±0.0009) & 0.1192 (±0.0004) & 19.0637 (±0.0491) & 0.1069 (±0.0004) \\
& U-NET & 0.7479 (±0.0006) & 0.5361 (±0.0005) & 92.2608 (±0.2457) & 0.4638 (±0.0005) \\
\hline
\multirow{5}{*}{50\%}
& Custom & 0.1865 (±0.0012) & \cellcolor{green!25}0.0432 (±0.0001) & 9.4832 (±0.0269) & \cellcolor{green!25}0.0391 (±0.0001) \\
& Augmented & \cellcolor{green!25}0.1850 (±0.0010) & \cellcolor{blue!15}0.0440 (±0.0001) & \cellcolor{green!25}9.3706 (±0.0292) & \cellcolor{blue!15}0.0394 (±0.0001) \\
& GP & \cellcolor{blue!15}0.1915 (±0.0004) & 0.0638 (±0.0003) & \cellcolor{blue!15}13.7339 (±0.0531) & 0.0602 (±0.0002) \\
& CSDI & 0.3913 (±0.0007) & 0.1239 (±0.0002) & 20.3463 (±0.0494) & 0.1115 (±0.0001) \\
& U-NET & 0.8144 (±0.0007) & 0.5857 (±0.0010) & 101.1553 (±0.3862) & 0.5098 (±0.0008) \\
\hline
\multirow{5}{*}{75\%}
& Custom & 0.2933 (±0.0009) & 0.0656 (±0.0001) & 13.2244 (±0.0311) & 0.0588 (±0.0002) \\
& Augmented & \cellcolor{green!25}0.2744 (±0.0018) & \cellcolor{green!25}0.0633 (±0.0002) & \cellcolor{green!25}12.0324 (±0.0540) & \cellcolor{green!25}0.0558 (±0.0003) \\
& GP & \cellcolor{blue!15}0.3839 (±0.0059) & \cellcolor{blue!15}0.1044 (±0.0009) & \cellcolor{blue!15}18.9232 (±0.0734) & \cellcolor{blue!15}0.1010 (±0.0006) \\
& CSDI & 0.4263 (±0.0014) & 0.1441 (±0.0004) & 23.6783 (±0.0289) & 0.1324 (±0.0003) \\
& U-NET & 0.9145 (±0.0006) & 0.6431 (±0.0004) & 111.1768 (±0.1814) & 0.5686 (±0.0003) \\
\hline
\multirow{5}{*}{90\%}
& Custom & 0.3563 (±0.0021) & 0.1127 (±0.0006) & 22.3026 (±0.1002) & 0.1030 (±0.0005) \\
& Augmented & \cellcolor{green!25}0.3212 (±0.0020) & \cellcolor{green!25}0.0957 (±0.0004) & \cellcolor{green!25}18.5258 (±0.0629) & \cellcolor{green!25}0.0852 (±0.0004) \\
& GP & \cellcolor{blue!15}0.5257 (±0.0012) & \cellcolor{blue!15}0.1596 (±0.0004) & \cellcolor{blue!15}26.2071 (±0.0546) & \cellcolor{blue!15}0.1563 (±0.0003) \\
& CSDI & 0.5608 (±0.0024) & 0.2360 (±0.0005) & 39.5478 (±0.0619) & 0.2194 (±0.0003) \\
& U-NET & 0.9850 (±0.0004) & 0.6862 (±0.0004) & 118.6330 (±0.1639) & 0.6157 (±0.0003) \\
\hline
\end{tabular}}
\caption*{(a) Imputation Metrics}
\end{minipage}

\hfill

\begin{minipage}{0.75\linewidth}
\centering
\resizebox{\linewidth}{!}{
\begin{tabular}{|l|l|c|c|c|c|}
\hline
Horizon & Model & RMSE & MAE & SMAPE & CRPS \\
\hline
\multirow{5}{*}{6}
& Custom & \cellcolor{blue!15}0.3211 (±0.0021) & \cellcolor{blue!15}0.0878 (±0.0010) & \cellcolor{blue!15}17.3480 (±0.1479) & \cellcolor{blue!15}0.0822 (±0.0009) \\
& Augmented & \cellcolor{green!25}0.2679 (±0.0013) & \cellcolor{green!25}0.0773 (±0.0002) & \cellcolor{green!25}15.3664 (±0.0752) & \cellcolor{green!25}0.0702 (±0.0003) \\
& CSDI & 0.3887 (±0.0024) & 0.1260 (±0.0003) & 22.1140 (±0.0881) & 0.1171 (±0.0005) \\
& GP & 0.4059 (±0.0072) & 0.1126 (±0.0015) & 20.2815 (±0.1756) & 0.1059 (±0.0007) \\
& U-NET & 0.9606 (±0.0012) & 0.6716 (±0.0010) & 116.9822 (±0.4429) & 0.6058 (±0.0009) \\
\hline
\multirow{5}{*}{12}
& Custom & \cellcolor{blue!15}0.3895 (±0.0007) & \cellcolor{blue!15}0.1140 (±0.0004) & \cellcolor{blue!15}21.1204 (±0.1045) & \cellcolor{blue!15}0.1066 (±0.0006) \\
& Augmented & \cellcolor{green!25}0.3013 (±0.0024) & \cellcolor{green!25}0.1038 (±0.0003) & \cellcolor{green!25}19.5852 (±0.0642) & \cellcolor{green!25}0.0936 (±0.0003) \\
& CSDI & 0.4351 (±0.0023) & 0.1489 (±0.0005) & 24.8425 (±0.0694) & 0.1420 (±0.0004) \\
& GP & 0.4870 (±0.0090) & 0.1618 (±0.0007) & 26.4087 (±0.0674) & 0.1540 (±0.0009) \\
& U-NET & 1.0037 (±0.0006) & 0.6962 (±0.0005) & 120.2760 (±0.1908) & 0.6297 (±0.0006) \\
\hline
\multirow{5}{*}{18}
& Custom & \cellcolor{blue!15}0.4526 (±0.0009) & \cellcolor{blue!15}0.1473 (±0.0004) & \cellcolor{blue!15}25.0509 (±0.0569) & \cellcolor{blue!15}0.1374 (±0.0007) \\
& Augmented & \cellcolor{green!25}0.3412 (±0.0017) & \cellcolor{green!25}0.1317 (±0.0004) & \cellcolor{green!25}23.3131 (±0.0377) & \cellcolor{green!25}0.1186 (±0.0004) \\
& CSDI & 0.5019 (±0.0007) & 0.1779 (±0.0004) & 27.0204 (±0.0746) & 0.1747 (±0.0003) \\
& GP & 0.5640 (±0.0046) & 0.2010 (±0.0009) & 30.3275 (±0.0780) & 0.1968 (±0.0008) \\
& U-NET & 1.0199 (±0.0003) & 0.7050 (±0.0004) & 121.3809 (±0.1661) & 0.6379 (±0.0003) \\
\hline
\end{tabular}}
\caption*{(b) Forecasting Metrics}
\end{minipage}
\caption{
OPE dataset performance comparison. Best values are highlighted in green; second-best in blue.
\textbf{Top}: Imputation metrics across missing ratios.
\textbf{Bottom}: Forecasting metrics across horizons.
}
\label{fig:all_results_real}
\vspace{-4mm}
\end{figure*}

\end{document}